\documentclass[letterpaper, 10 pt, conference]{ieeeconf}  

\IEEEoverridecommandlockouts                              

\usepackage{graphicx} 
\usepackage{subcaption}
\usepackage{amssymb,amsfonts,amsmath,amscd}
\usepackage{units}
\usepackage[]{algorithm2e}
\usepackage{booktabs}
\usepackage[export]{adjustbox}

\title{\LARGE \bf
Mapless Online Detection of Dynamic Objects in 3D Lidar
}

\author{David J. Yoon, Tim Y. Tang, and Timothy D. Barfoot
\thanks{The authors are with the University of Toronto Institute for Aerospace Studies, Toronto, Ontario, Canada.
        {\tt\small \{david.yoon, \newline tim.tang\}@robotics.utias.utoronto.ca}, {\tt\small tim.barfoot@utoronto.ca}}
}

\begin{document}

\maketitle
\thispagestyle{empty}
\pagestyle{empty}

\begin{abstract}
This paper presents a model-free, setting-independent method for online detection of dynamic objects in 3D lidar data. We explicitly compensate for the moving-while-scanning operation (motion distortion) of present-day 3D spinning lidar sensors. Our detection method uses a motion-compensated freespace querying algorithm and classifies between dynamic (currently moving) and static (currently stationary) labels at the point level. For a quantitative analysis, we establish a benchmark with motion-distorted lidar data using CARLA, an open-source simulator for autonomous driving research. We also provide a qualitative analysis with real data using a Velodyne HDL-64E in driving scenarios. Compared to existing 3D lidar methods that are model-free, our method is unique because of its setting independence and compensation for pointcloud motion distortion.
\end{abstract}

\section{INTRODUCTION}

An autonomous system must be aware of dynamic elements in its environment. Our focus is on detection using lidar (light detection and ranging), specifically spinning lidars, which operate by sweeping multiple lasers about an axis for a $360^\circ$ field of view (FOV). For the remainder of this paper, we refer to detectable elements as objects.

We place detection methods into three categories: methods that use class-specific detectors \cite{petrov_au09} \cite{chen_cvpr17}, methods that use maps \cite{hebel2011change} \cite{underwood_icra13} \cite{pomerleau_icra14}, and methods that only use recently acquired data (live data) \cite{dewan_icra16} \cite{dewan_iros16}. Class-specific, or model-based, detectors use prior information of objects and therefore have restricted use. Methods that use maps are also restricted, but may not require prior information on objects, which we refer to as model-free. Lastly, methods that only use live data may not require prior information on objects or the setting (e.g., maps), at the cost of being unable to detect stationary objects that have the potential to move (e.g., stationary cars in traffic). The categories are complementary, and therefore a combination can be effective \cite{moosmann_icra13} \cite{ushani_icra17} \cite{dewan_iros17}.

\begin{figure}[ht]
\vspace*{0.2cm}
  \centering
    \includegraphics[width=0.48\textwidth]{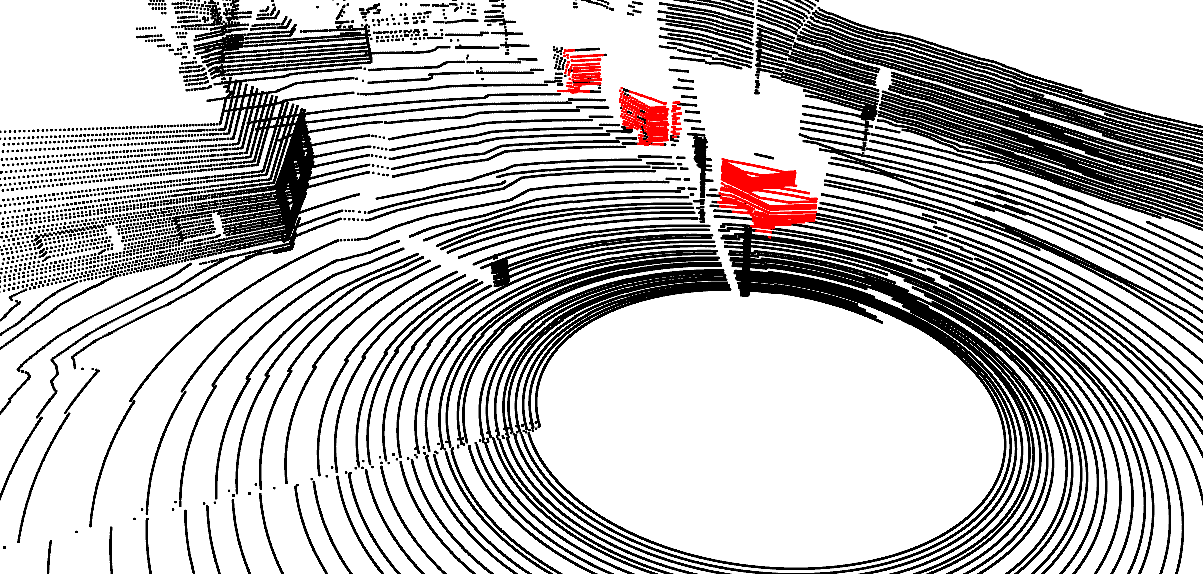}
  \caption{Example pointcloud of simulated lidar data (CARLA \cite{carla}). The output of our dynamic object detection method is indicated by the coloured points. We make our motion-distorted datasets with point-level groundtruth on dynamic objects available as a public benchmark.}
  \label{sim}
  \vspace*{-0.6cm}
\end{figure}

Our work belongs to the category that only uses live data. We detect objects, regardless of the class or setting, as long as they are moving in the current scene. Our method is therefore applicable to a large variety of applications. In the urban driving setting, an application dominated by deep learning methods, our model-free detector could be used as a safety net. In a disaster zone scenario, where a prior map does not exist because of dramatic changes to the scene, our detector could be used to identify survivors. Our method is isotropic to the setting (e.g., we do not exploit knowledge of the gravity vector to help detect a ground plane), which is a contributing factor to setting-independence.


Our main contribution is setting-independent detection that outputs point-level labels as dynamic (currently moving) or static (currently stationary). We use a lidar odometry algorithm from previous work that compensates for motion distortion caused by the moving-while-scanning operation of spinning lidars \cite{tang_crv18} \cite{mcgarey_fsr17}. We use a novel freespace querying algorithm that also compensates for motion distortion. To the best of our knowledge, our method is the only existing among other model-free ones that combines all of the following traits: motion-compensation, environment isotropy, only uses live data (i.e., no maps or training data). 

Our secondary contribution is a benchmark of simulated data using CARLA \cite{carla}, an open-source simulator for autonomous driving research (see example in Fig. \ref{sim}), which we believe to be the first to provide motion-distorted data with per-point groundtruth labels on moving objects. We also provide a qualitative evaluation of our pipeline with real data collected with a Velodyne HDL-64E sensor on a vehicle.

The remainder of the paper is organized as follows. Section \ref{related_work} discusses literature; Section \ref{methodology} describes the pipeline methodology; Section \ref{results} presents the simulation, experiment, and performance analysis; and Section \ref{conclusions} provides concluding thoughts and future work recommondations. 

\section{RELATED WORK} \label{related_work}

Class-specific, or model-based, detectors take advantage of prior information of the objects to be detected. Petrovskaya et al. \cite{petrov_au09} model vehicles as 2D bounding boxes, which is applied to 3D data by processing it into a 2D representation. Rather than manually crafting models, recent work focuses on learning methods. Chen et al. \cite{chen_cvpr17} (among many others) take lidar and camera data as input to a deep neural network (DNN) and output class-specific detections. While this category is proven to work well, such methods will simply not detect objects for which they have not been trained.

Detection without prior object information is possible by comparing current data to a reliable prior map. Given a reliable map of the stationary world, differences from the comparison are indicative of dynamic objects. Sometimes called \textit{change detection} \cite{hebel2011change} \cite{underwood_icra13}, these methods make use of pointcloud comparisons (i.e., end-points of lidar measurements) and freespace comparisons (i.e., paths traced by lidar measurements). Hebel et al. \cite{hebel2011change} raytrace lidar data into occupancy voxel grids for their freespace representation. Occupancy voxel grids are expensive computationally and in memory, so instead, Pomerleau et al. \cite{pomerleau_icra14} query lidar freespace by matching measurements with local spherical coordinates. The method is efficient, but assumes pointclouds are not motion distorted. Note that while existing works that use occupancy voxel grids do not consider motion distortion, compensation is trivial with a continuous-time trajectory.

Our interest is in methods that do not require prior information on the objects or the setting, only making use of the latest lidar data (live data). Such methods are limited to detecting objects that are moving in the current scene. Objects that are stationary, but may be of interest (e.g., stationary cars in traffic), are not detectable by such methods.

Among live data methods are ones that only use pointcloud information. Dewan et al. \cite{dewan_icra16} compare subsequent pointclouds and sequentially identify motion through a voting scheme. The first detected motion will always be the relative motion of the stationary environment, followed by the largest dynamic objects. They directly compared their work to Moosmann and Stiller \cite{moosmann_icra13} and showed superior performance at the object level. In another publication, Dewan et al. \cite{dewan_iros16} produce scene flow (i.e., point-wise velocity estimation), from which dynamic labels are trivial. However, both of their methods are not setting-independent because they remove ground points as a pre-processing step. Live data methods are challenging because there are significant differences between subsequent pointcloud comparisons due to viewpoint occlusions or data sparsity. Removing ground points is helpful in avoiding false detections, but it is not clear how to handle occlusions with only pointclouds.

Live data methods that use freespace handle viewpoint occlusions well. Azim and Aycard \cite{azim_iv12} raytrace over occupancy voxel grids and compare them over time, but only provide a qualitative analysis of their method. Postica et al. \cite{postica_iros16} also make comparisons with occupancy voxel grids. They present quantitative results using short sequences from the KITTI dataset \cite{Geiger2013IJRR}, for which they have manually annotated for groundtruth, but have not made public. Notable limitations of their work include relying on pre-processing ground points and ignoring measurements further than $30$ m.

The three categories are complementary to one another, so a combination can be more effective. Moosmann and Stiller \cite{moosmann_icra13} segment pointclouds into object proposals, which they track over time. Consistent ones are labelled dynamic by a learned classifer. Ushani et al. \cite{ushani_icra17} use freespace and learning to compute scene flow. Occupancy voxel grids coarsely identify dynamic points, which are then refined by a learned classifier. They make a planar motion assumption and limit their method to a 50 m $\times$ 50 m grid. Dewan et al. combine their prior work on scene flow \cite{dewan_iros16} with a DNN to produce point labels of dynamic, static, and a third label for objects with the potential to move (e.g., stationary cars) \cite{dewan_iros17}.

Our method belongs to the live data category and is model-free, labelling each point as dynamic or static. We motion-compensate both pointclouds and freespace querying, which existing methods do not consider. We make full use of the sensor range, which is often limited for methods that use freespace. Since our method only uses live data and is isotropic, it is setting-independent. If setting independence is not important, our method can be combined with strategies from the other categories for greater performance.

\section{METHODOLOGY} \label{methodology}
\begin{figure*}[ht]
  \vspace*{0.2cm}
  \centering
    \includegraphics[width=1\textwidth]{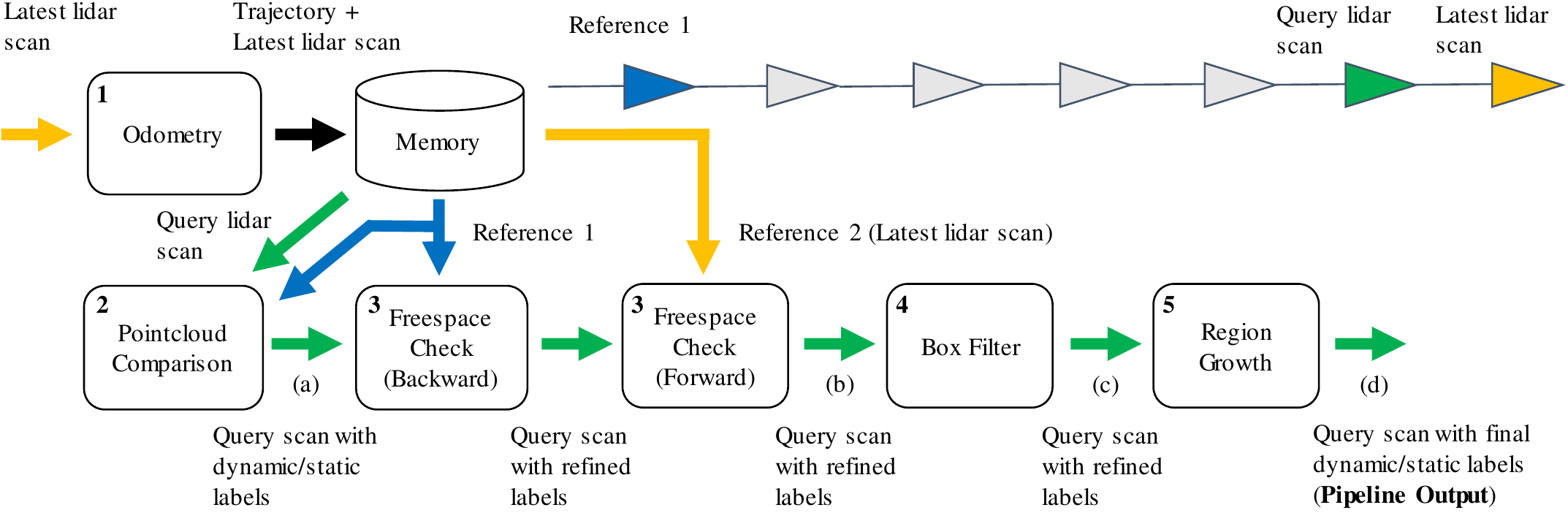}
  \caption{The pipeline describes the sequence of operations on the lidar scan of interest (\textit{query scan}), outputing the scan with points labelled dynamic or static. A lidar odometry algorithm computes the sensor trajectory, which aligns the \textit{latest scan}. The labels (a) to (d) correspond to the images in Fig. \ref{cascade}. The numbers correspond to the enumerated steps in the text.}
  \label{pipeline}
  \vspace*{-0.6cm}
\end{figure*}

\begin{figure}
    \centering
    \begin{subfigure}[b]{0.235\textwidth}
        \centering
        \includegraphics[width=\textwidth]{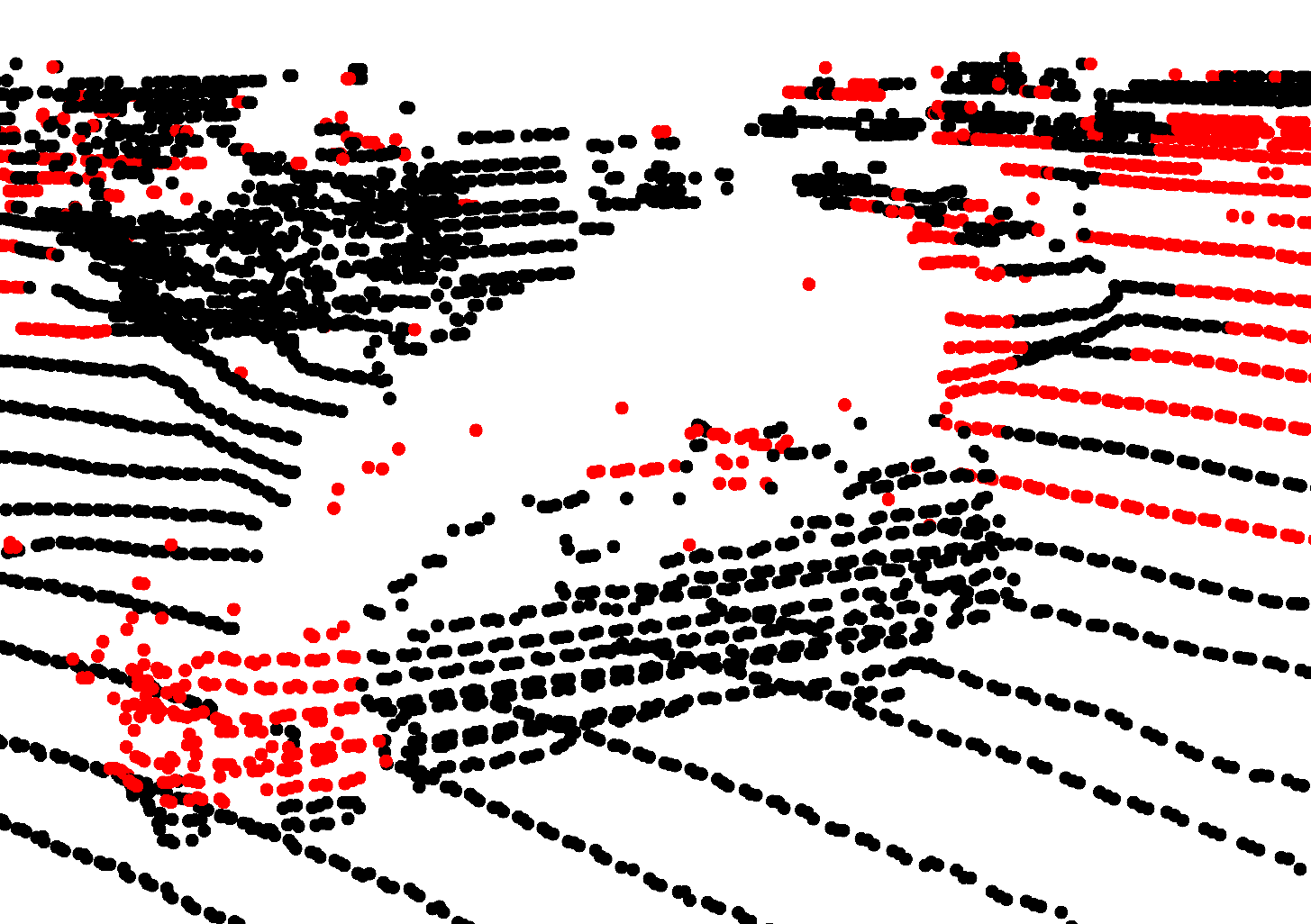}
        \caption[After pointcloud comparison.]%
        {{\small After pointcloud comparison.}}    
        \label{fig:mean and std of net14}
    \end{subfigure}
    \hfill
    \begin{subfigure}[b]{0.235\textwidth}  
        \centering 
        \includegraphics[width=\textwidth]{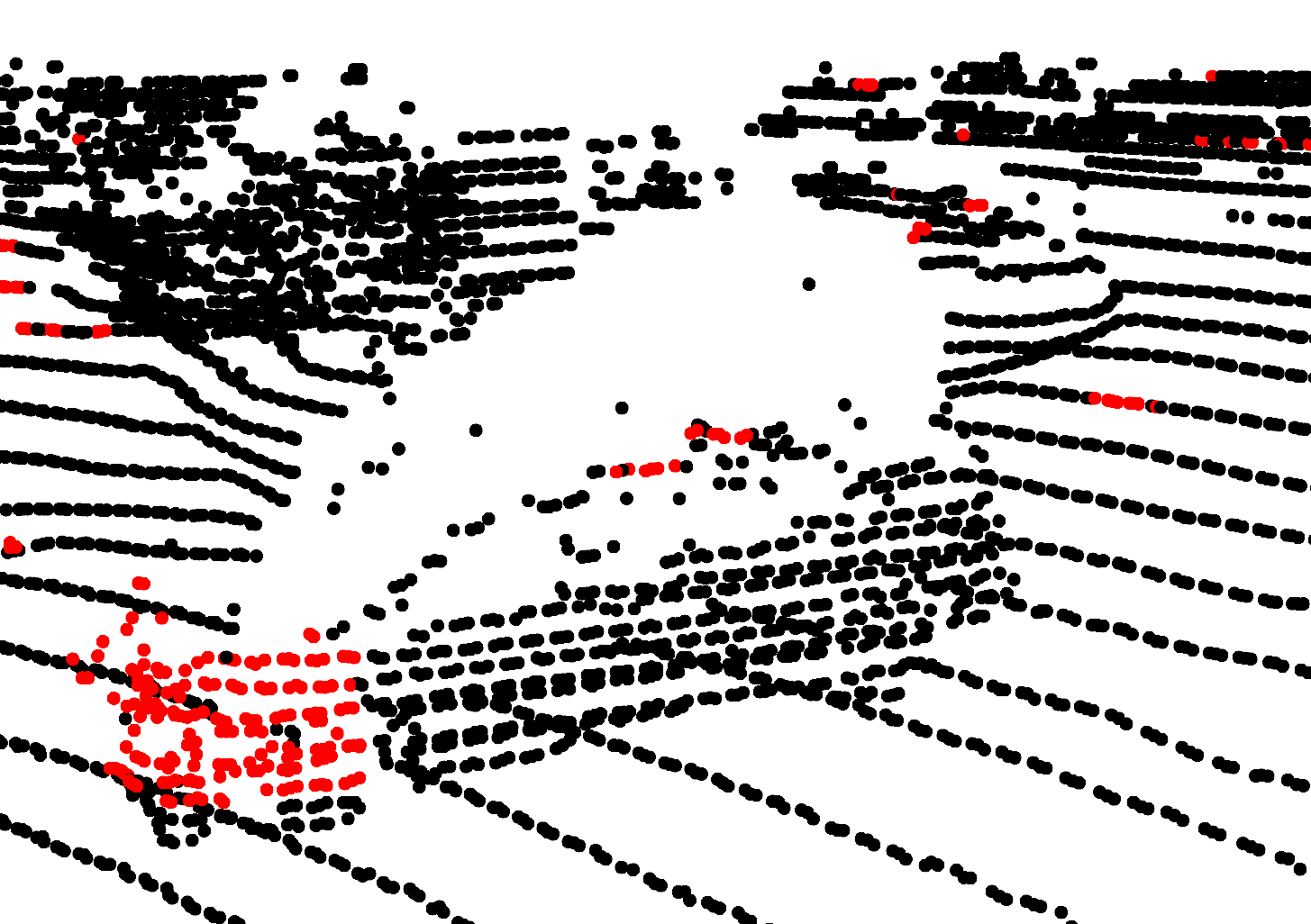}
        \caption[After freespace check.]%
        {{\small After freespace check.}}    
        \label{fig:mean and std of net24}
    \end{subfigure}
    \begin{subfigure}[b]{0.235\textwidth}   
        \centering 
        \includegraphics[width=\textwidth]{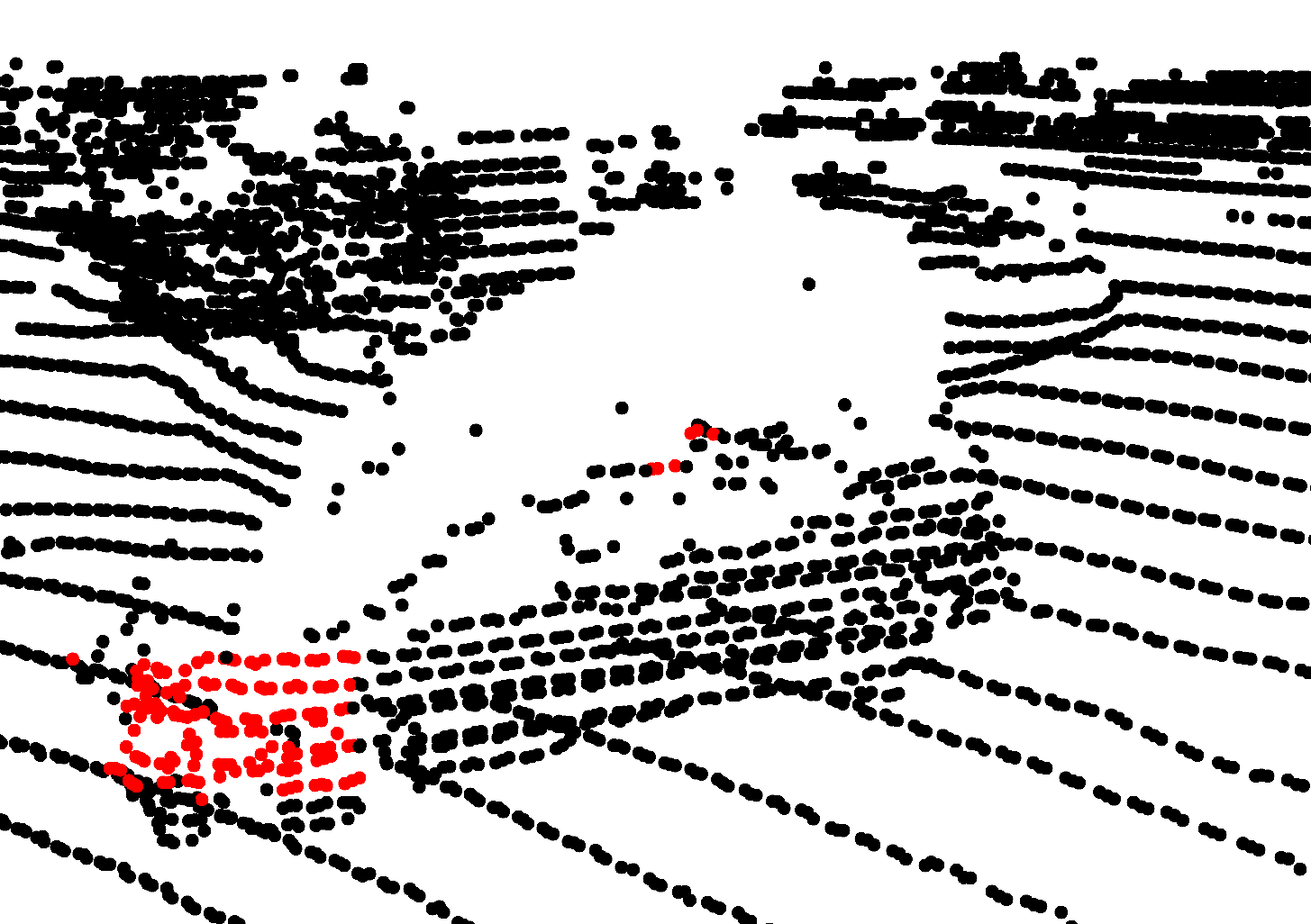}
        \caption[After box filter.]%
        {{\small After box filter.}}    
        \label{fig:mean and std of net34}
    \end{subfigure}
    \hfill
    \begin{subfigure}[b]{0.235\textwidth}   
        \centering 
        \includegraphics[width=\textwidth]{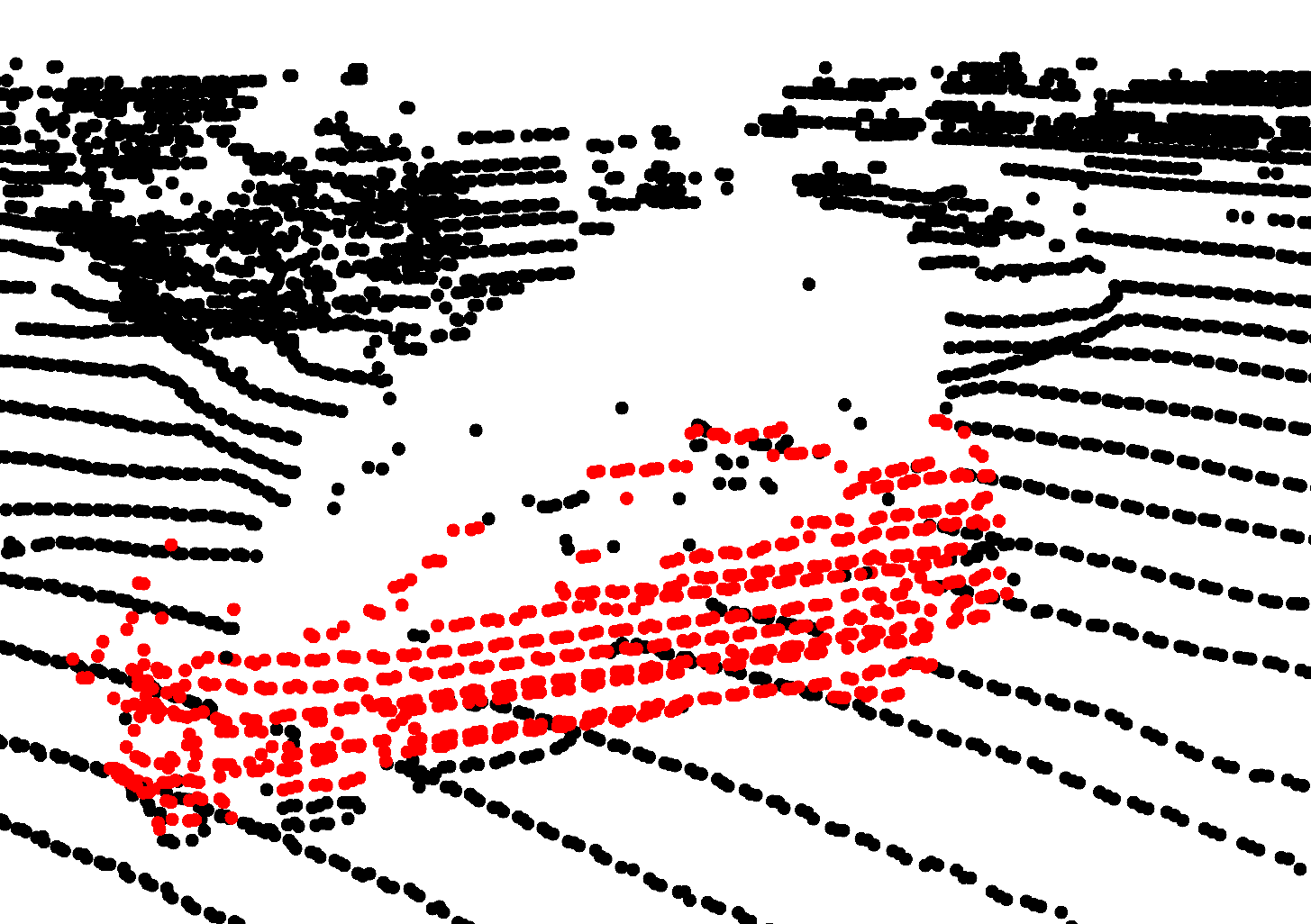}
        \caption[After region growth.]%
        {{\small After region growth.}}    
        \label{fig:mean and std of net44}
    \end{subfigure}
    \caption[]
    {A vehicle throughout the detection pipeline. Refer to the pipeline in Fig. \ref{pipeline} for corresponding letters (a) to (d).} 
    \label{cascade}
    \vspace*{-0.5cm}
\end{figure}
We are interested in labelling lidar points as static or dynamic. We define a lidar scan as a full revolution of the spinning lidar which, depending on the lidar, can have more than 100000 points. Each laser of a spinning lidar has a unique position and orientation on the rotating base. This means the laser ray paths (i.e., freespace) are not equivalent to azimuth and elevation computations of the endpoints, which are defined in a local reference frame. Thus we refer to scans as including both pointcloud and freespace information. We refer to measurements as points when appropriate. In a full pass of the pipeline, we label all points in a single scan of interest (\textit{query scan}).

A pipeline diagram is shown in Fig. \ref{pipeline}. The sequential steps to the pipeline are explained as follows:
\begin{enumerate}
  \item \textit{Odometry}: Align \textit{latest scan} using lidar odometry.
  \item \textit{Pointcloud Comparison}: Comparison of query scan against another. Discrepencies are set to dynamic.
  \item \textit{Freespace Check}: Check dynamic points against freespace of another scan. Points not in freespace are not dynamic and changed to static.
  \item \textit{Box Filter}: Apply a sliding box filter on the image representation of the query scan for outlier rejection.
  \item \textit{Region Growth}: Cluster the dynamic query scan points. Add nearby points to clusters if they satisfy conditions that indicate they are part of the same object.
\end{enumerate}
We spend the rest of this section to discuss each step in more detail and highlight its contribution to the pipeline.

\subsection{Odometry}
Our lidar odometry considers the exact timestamps of measurements to produce a continuous-time trajectory of the moving sensor platform. For detail, please see our previous work \cite{tang_crv18}, which uses the trajectory representation of Anderson and Barfoot \cite{anderson_iros15}.

With a continuous-time trajectory, we can query for any pose and velocity given a timestamp, $t$. We use the notation $\mathbf{T}_{v,0}\left(t \right) \in SE(3)$, where $\mathbf{T}_{v,0}$ is a transformation from the world frame, $\underrightarrow{\boldsymbol{\mathcal{F}}}_0$, to the moving sensor platform, $\underrightarrow{\boldsymbol{\mathcal{F}}}_v$. $\boldsymbol{\varpi}(t)$ is the body-centric velocity of the sensor platform.

\subsection{Pointcloud Comparison} \label{pointcloud_comparison}
We compute a pointcloud comparison between the query scan and a previous \textit{reference scan}. The scans are of different time intervals, so dynamic objects cause discrepencies.

We identify discrepencies using error metrics commonly used in pointcloud alignment problems. We use a point-to-plane metric when points have sufficient neighbours to compute surface normals. Otherwise, we use a point-to-point metric. All points are transformed to $\underrightarrow{\boldsymbol{\mathcal{F}}}_0$ using $\mathbf{T}_{v,0}\left(t \right)$, creating motion-compensated pointclouds. Given a \textit{query point}, $\mathbf{q}_0$, its normalized surface normal, $\mathbf{n}^q$, and its nearest reference scan neighbour, $\mathbf{p}_0$, the point-to-plane metric is $\lvert \mathbf{n}^q \cdot \left( \mathbf{p}_0 - \mathbf{q}_0 \right) \rvert$. The point-to-point metric is $\| \mathbf{p}_0 - \mathbf{q}_0 \|_2$.

We compute the error metric for all query points. Static points have low error because there should be a corresponding reference scan point of the same surface observation. We expect high error from dynamic points since they are observations of moving surfaces. However, a moving $\underrightarrow{\boldsymbol{\mathcal{F}}}_v$ causes new surface observations and viewpoint occlusions, which also cause high error. Regardless, we take a constant scalar \textit{error threshold}. Those greater than the error threshold are labelled dynamic, the rest are static (see Fig. \ref{cascade}a). We refine incorrect dynamic labels later in the pipeline. 

Notice there is a \textit{scan gap} of $4$ between the query scan and reference scan (Reference 1) in the top-right illustration of Fig. \ref{pipeline}. We require a scan gap to ensure dynamic objects sufficiently displace between the two scans. Fig. \ref{cascade}a shows a partially labelled object since the ideal scan gap requires unknown quantities (i.e., object speed and size). Instead, we aim to partially label objects and recover the rest later in the pipeline. The error threshold, scan gap, and lidar scan rate together define the minimum speed of detectable objects.

\subsection{Freespace Check}
We check all dynamic query points, which can be as many as half the query scan points, against the freespace of another scan to correct mislabels from the pointcloud comparison. Recall that points are mislabelled dynamic because of viewpoint occlusions or they are new surface observations. Dynamic points inside freespace are consistent with their current label, while ones on the border or outside freespace may not truly be dynamic. We use this argument to refine incorrect dynamic labels (see Fig. \ref{cascade}a and Fig. \ref{cascade}b). We do not check freespace for static points since a pointcloud comparison is equivalent to a freespace border check.

Given a query point and reference scan that defines the freespace of interest, we wish to determine if the query point is inside, on the border of, or outside freespace. Recall that the spinning lidar continuously sweeps lasers about an axis for a $360^\circ$ FOV. The laser ray paths, from the sensor to their endpoints, define freespace. We designed our freespace method specifically for a single lidar configuration that has its lasers approximately radiate outward vertically (i.e., along the sweeping axis). This is important because we exploit the elevation order of the lasers to speed up our freespace query.

Representing freespace in its entirety is possible, but expensive (e.g., occupancy voxels). Instead, we only determine the reference scan measurement ray that has a direction that passes nearest to the query point (i.e., smallest point-to-line distance). Consider the query point surface plane and the identified reference scan ray -- there are three cases:
\begin{itemize}
    \item Case 1: Ray intersects the surface plane.
    \item Case 2: Ray lies on the surface plane.
    \item Case 3: Ray does not reach the surface plane.
\end{itemize}
Case 1 means the surface plane is absent during the time period of the reference scan, which is possible if the query point is dynamic and inside freespace. Case 2 means the measurement is an observation of the same surface plane (freespace border). Finally, Case 3 means another surface obstructed the measurement (outside freespace). 

Pomerleau et al. \cite{pomerleau_icra14} also use a nearest-ray strategy by searching a kd-tree of spherical coordinates, but assume their pointclouds are instantaneous snapshots (ideal pointcloud) of the scene. Instead, we compensate for the sensor platform motion by using our continuous-time trajectory, $\mathbf{T}_{v,0}(t)$.

We assume a total of $L$ lasers rotate together at constant speed, $\omega$. Each laser $\ell$, indexed by increasing elevation, has a unique pose with respect to the sensor hub (i.e., the rotating base), $\underrightarrow{\boldsymbol{\mathcal{F}}}_h$, defined by the transformation $\mathbf{T}_{\ell,h} \in SE(3)$. 

\begin{figure}[h]
\vspace*{-0.3cm}
  \centering
    \includegraphics[width=0.4\textwidth]{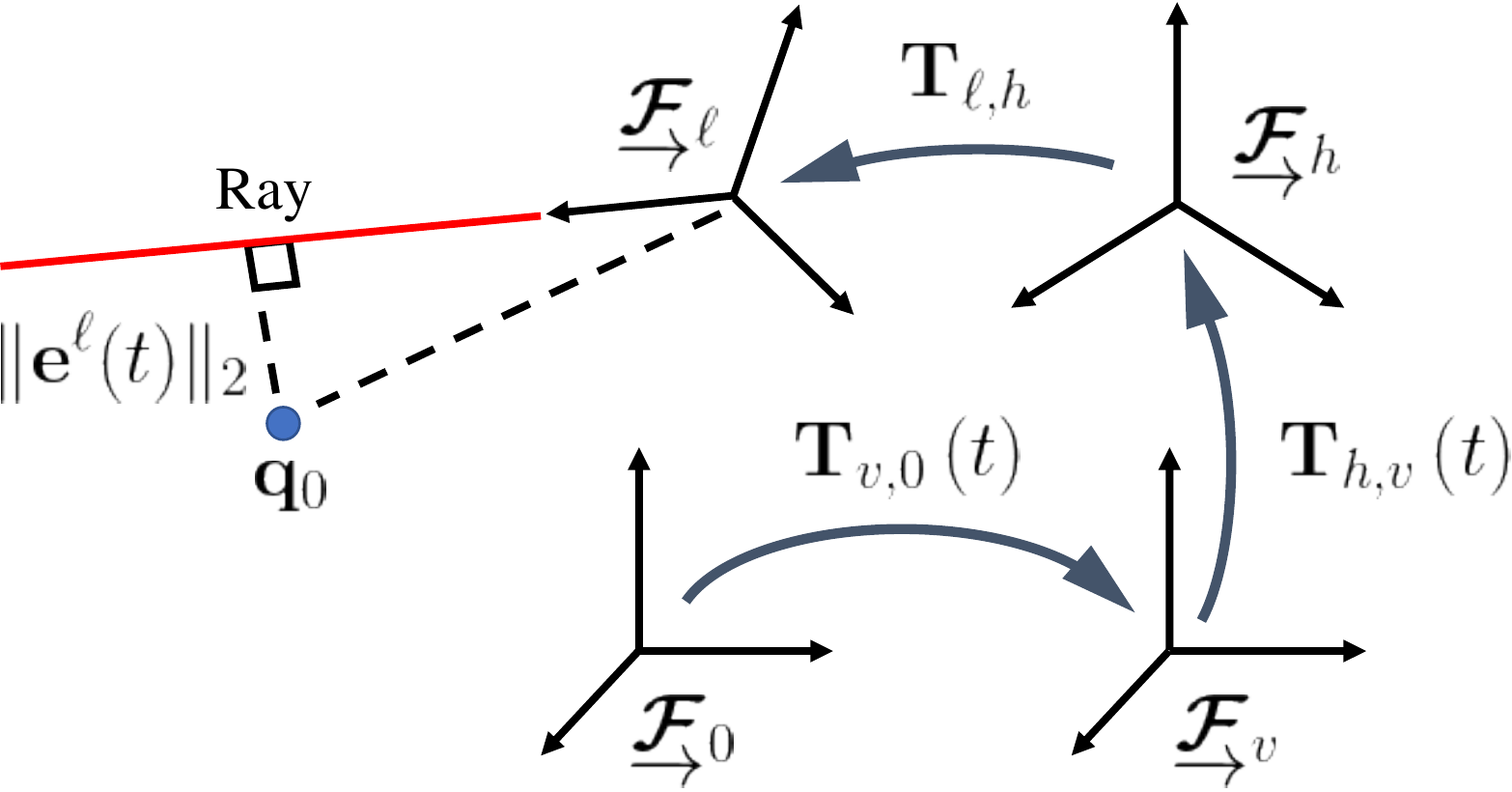}
  \caption{A visualization of the point-to-line distance (see Eq. \ref{erroreq}). $\mathbf{T}_{v,0}$ is the sensor trajectory, $\mathbf{T}_{h,v}$ is the spinning lidar rotation, and $\mathbf{T}_{\ell,h}$ is the unique pose for laser $\ell$. We require the time, $t$, and laser, $\ell$, combination that minimizes $\| \mathbf{e}^{\ell}(t) \|_2$.}
  \label{transform}
  \vspace*{-0.2cm}
\end{figure}

Given a query point, $\mathbf{q}_0$, we formulate for each laser, $\ell$, the point-to-line distance as a continuous function of time: 
\begin{equation}
  \| \mathbf{e}^{\ell}\left(t\right) \|_2 = \left\| \mathbf{D} \mathbf{T}_{\ell,h} \mathbf{T}_{h,v} \left( t \right) \mathbf{T}_{v,0} \left( t \right) \begin{bmatrix} \mathbf{q}_0 \\ 1 \end{bmatrix} \right\|_2.
  \label{erroreq}
\end{equation}
\begin{equation*}
  \mathbf{T}_{h,v} \left(t \right) = \begin{bmatrix} \mathbf{R}^z(\omega t) & \mathbf{0} \\ \mathbf{0}^T & 1 \end{bmatrix} \in SE(3), \quad \mathbf{D} = \begin{bmatrix} 0 & 0 & 0 & 0 \\ 0 & 1 & 0 & 0 \\ 0 & 0 & 1 & 0 \end{bmatrix},
\end{equation*}
where $\mathbf{R}^z(\omega t) \in SO(3)$ is the constant rotation of $\underrightarrow{\boldsymbol{\mathcal{F}}}_h$ with respect to $\underrightarrow{\boldsymbol{\mathcal{F}}}_v$ at rotation speed $\omega$. Note we define $\underrightarrow{\boldsymbol{\mathcal{F}}}_v$ such that there is no translation between $\underrightarrow{\boldsymbol{\mathcal{F}}}_h$ and $\underrightarrow{\boldsymbol{\mathcal{F}}}_v$, and $\underrightarrow{\boldsymbol{\mathcal{F}}}_h$ rotates about the $z$-axis of $\underrightarrow{\boldsymbol{\mathcal{F}}}_v$. We define $\underrightarrow{\boldsymbol{\mathcal{F}}}_{\ell}$ such that the laser points along the $x$-axis. See Fig. \ref{transform} for a visualization.

We require $t^*$ and $\ell^*$ that minimize $\| \mathbf{e}^{\ell}\left(t\right) \|_2$. Unfortunately, $\ell$ is discrete. We minimize $\| \mathbf{e}^{\ell}\left(t\right) \|_2$ iteratively by selecting $\ell$ and solving for $t$ using least-squares optimization:
\begin{equation}
t = 
  \begin{aligned}
  & \text{arg}\underset{t}{\text{\,min}}
  \,\, \frac{1}{2} \, \mathbf{e}^{\ell}(t)^T \mathbf{e}^{\ell}(t).
  \end{aligned}
  \label{optimization}
\end{equation}
We iterate by exploiting laser elevation order. We first solve Eq. \ref{optimization} using initial guesses for $t$ and $\ell$, the laser neighbour above it, $\ell + 1$, and below it, $\ell - 1$. If a neighbour optimizes to a smaller $\| \mathbf{e}^{\ell}\left(t\right) \|_2$, we iteratively search along that direction by single laser increments, re-solving Eq. \ref{optimization} and comparing optimized $\| \mathbf{e}^{\ell}\left(t\right) \|_2$ values. The iteration stops once $\| \mathbf{e}^{\ell}\left(t\right) \|_2$ no longer decreases or we run out of neighbours.

Our iterative method relies on a good initial condition for $t$ and $\ell$, which we select by using the ideal pointcloud method of Pomerleau et al. \cite{pomerleau_icra14}. We experimentally verified this initialization choice always converges. 

Given $t^*$ and $\ell^*$, the reference scan ray is the one with the closest timestamp. Adhering to the three possible cases, we check for an intersection between the ray and surface plane of the query point. This is easily done by computing the point-to-plane error metric of Section \ref{pointcloud_comparison} with the ray endpoint for a freespace border test. Otherwise, the query point is inside or outside freespace depending on which side of its surface the endpoint resides in.

Fig. \ref{pipeline} shows two freespace blocks, \textit{backward} (comparison with a previous scan), followed by \textit{forward} (comparison with a later scan). Ideally, only backward is required. Unfortunately, objects moving away from the sensor will never be within freespace of a previous scan. Backward shares the same reference scan (or scan gap) as in the pointcloud comparison. A scan gap is not needed for forward because objects moving away have surface geometry perpendicular to the movement direction. We only need to compute forward for points identified as outside freespace in backward.

\subsection{Box Filter}
\begin{figure}[]
  \vspace*{0.2cm}
  \centering
    \fbox{\includegraphics[trim={2cm 0.5cm 1cm 8.6cm},clip,valign = c]{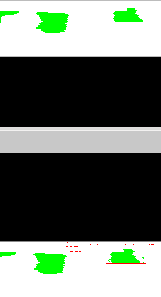}}
    $\begin{bmatrix}
      0 & 0 & 0 & 0 \\
      1 & 1 & 1 & 1 \\
      0 & 0 & 0 & 0
    \end{bmatrix}$
    \fbox{\includegraphics[trim={2cm 9.0cm 1cm 0.1cm},clip,valign = c]{fig/lidar_image3.png}}
  \caption{We apply a filter (middle) to a scan's image representation to remove mislabelled dynamic points (red) while maintaining correct ones (green). We show a before (left) and after (right) example (64 lasers, limited horizontal FOV).}
  \label{filter_image}
  \vspace*{-0.6cm}
\end{figure}

Our freespace check is susceptible to error because of finite lidar resolution (e.g., consider freespace at far ranges), leaving sparse traces of mislabelled dynamic points (see Fig. \ref{cascade}b). The query scan measurements are arranged into an image representation. Each laser forms a row and the consecutive measurements the columns. Dynamic labels have an image value of $1$ and static labels have a value of 0. We filter outliers (dynamic mislabels) by sliding a box filter thoughout the image (see Fig. \ref{filter_image} for an example).

We apply our filter (Fig. \ref{filter_image} middle) with a pixelwise XNOR (exclusive logical NOR) operation. The sum of all XNOR operations is a numerical score. Scores greater than a constant \textit{score threshold} are considered outliers. The score threshold depends on the lidar resolution.

\subsection{Region Growth}
Often, the scan gap is not large enough for an object to completely displace from its previous position (see Fig. \ref{cascade}c). Thus we require a method for region growing (see Fig. \ref{cascade}d).

We first cluster the dynamic query points into object clusters. We use the 3D pointcloud clustering method presented by Klasing et al. \cite{klassing_icra08}. A radially bounded nearest neighbour strategy incrementally groups dynamic points into clusters.

We region grow clusters using the iterative pointcloud segmentation method of Moosmann et al. \cite{moosmann_icra09}. Clusters are grown by testing neighbouring points for parallelism or convexity, until none are found. Given two points $\mathbf{p}^1$ and $\mathbf{p}^2$, with unit surface normals $\mathbf{n}^1$ and $\mathbf{n}^2$, the two points are convex if the following two conditions are both true:
\begin{equation*}
  \vspace*{-0.1cm}
  \mathbf{n}^1 \cdot \left( \mathbf{p}^2 - \mathbf{p}^1 \right) \leq 0, \quad \mathbf{n}^2 \cdot \left( \mathbf{p}^1 - \mathbf{p}^2 \right) \leq 0.
\end{equation*}

\section{RESULTS} \label{results}
\subsection{Simulated Lidar Benchmark (CARLA)}
We chose to create a simulated dataset because of the difficulty in obtaining accurate groundtruth labels at the point level. The KITTI Vision Benchmark Suite \cite{Geiger2013IJRR}, a popular dataset, lacks point-level labels and does not provide raw lidar data (range and bearing with laser positions and orientations). The Paris-Lille-3D lidar dataset \cite{roynard_ijrr18} has point-level labels, but similar to KITTI, only provides motion-compensated pointclouds. They also removed points with range further than 20 m. Publications with work comparable to ours either omit quantitative results or use short sequences of manually labelled data, which they do not make public.

CARLA is an open-source simulator for autonomous driving research \cite{carla}. Two urban scenarios are provided with 2.9 km (Town 1) and 1.9 km (Town 2) of drivable roads. We currently made 5 min sequences of each, which we plan on expanding in the near future (visit http://asrl.utias.utoronto.ca/datasets/mdlidar/index.html).

We encourage use of our dataset for comparison. We made modifications to the CARLA source code to produce datasets matching a real Velodyne HDL-64E (e.g., laser positions and orientations). We capture motion distortion by making each laser activate once every simulation step, resulting in 128000 measurements (120 m maximum range) at 10 Hz frequency. See an example pointcloud in Fig. \ref{sim}.

For groundtruth, points moving faster than 0.2 m/s are dynamic. True positives (TP) are points correctly labelled dynamic. False positives (FP) and False negatives (FN) are points incorrectly labelled dynamic and static, respectively.

We compute precision, P, and recall, R, in two ways. Given the scan index, $n$, and the total number of scans, $N$, the \textit{total} computation is:
\begin{equation}
\vspace*{-0.1cm}
  \text{P}_\textit{t} = \frac{ \sum_n^N \text{TP}_n }{ \sum_n^N \left( \text{TP}_n + \text{FP}_n \right) }, \, \text{R}_\textit{t} =  \frac{ \sum_n^N \text{TP}_n }{ \sum_n^N \left( \text{TP}_n + \text{FN}_n \right) }.
  \label{total_pr}
\end{equation}

Given the total number of valid scans for P, $N_p$, and the total number of valid scans for R, $N_r$, the \textit{average} computation is:
\begin{equation}
\vspace*{-0.1cm}
  \text{P}_\textit{a} = \frac{ \sum_n^{N_p} \nicefrac{\text{TP}_n}{\left(\text{TP}_n + \text{FP}_n\right)} }{ N_p }, \, \text{R}_\textit{a} = \frac{ \sum_n^{N_r} \nicefrac{\text{TP}_n}{\left(\text{TP}_n + \text{FN}_n\right)} }{ N_r }.
  \label{average_pr}
\end{equation}
Scans where the denominator is 0 are ignored (e.g., $\text{TP}_n + \text{FP}_n = 0$), which is why we distinguish $N_p$ and $N_r$.

\begin{figure}[ht]
  \vspace*{-0.2cm}
  \centering
    \includegraphics[width=0.5\textwidth]{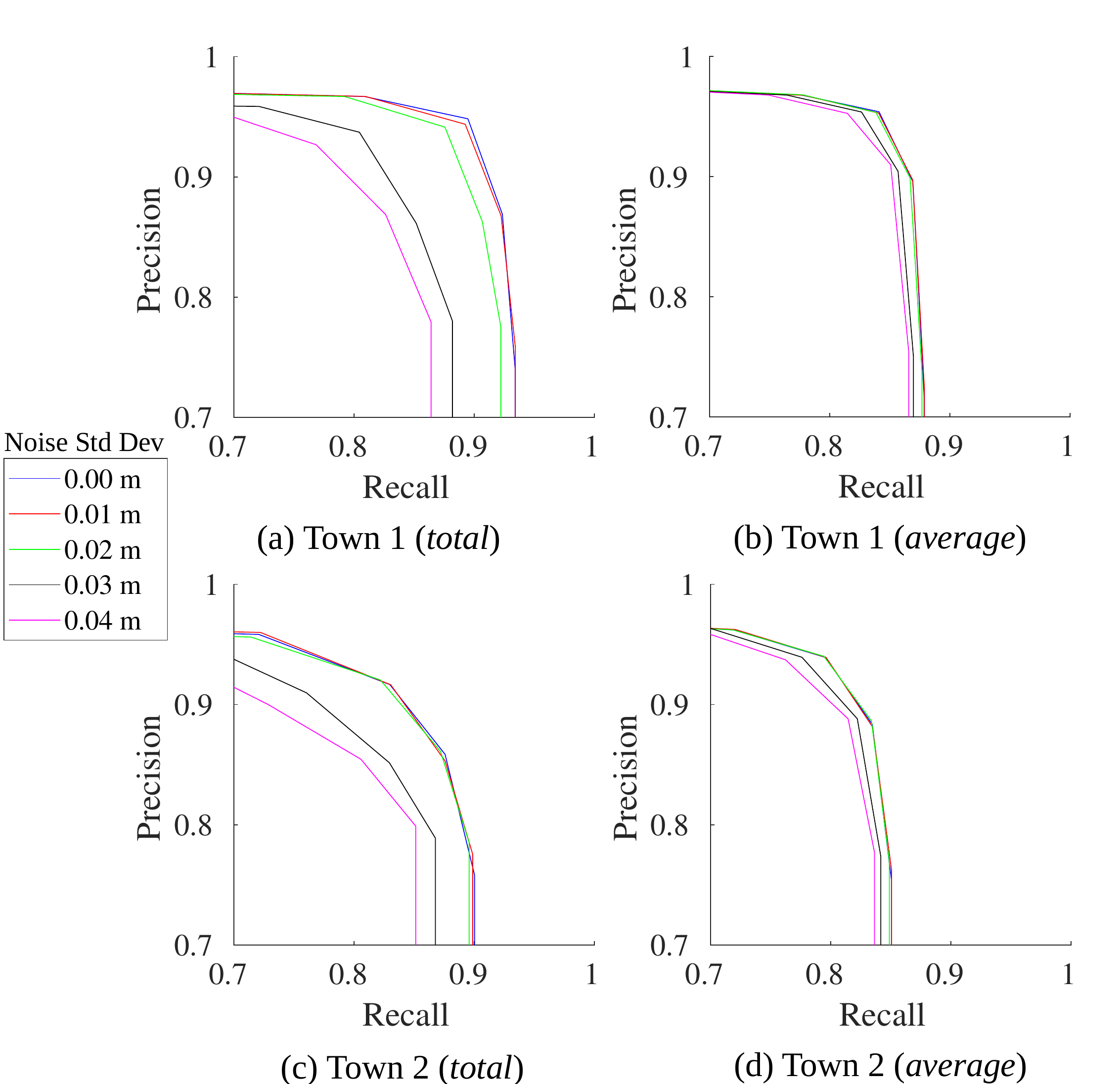}
  \caption{Precision-recall plots (\textit{total} and \textit{average} - see Eq. \ref{total_pr} and \ref{average_pr}) on two simulated sequences. The standard deviation of range measurement noise was varied. Note that a Velodyne HDL-64E has a standard deviation rated less than 0.02 m.}
  \label{pr}
  \vspace*{-0.2cm}
\end{figure}

\subsection{Benchmark Results}
We computed PR curves by varying the error threshold (see Fig. \ref{pr}). Noise was added to range measurements with varying standard deviation. A Velodyne HDL-64E has a range standard deviation rated less than 0.02 m. The scan gap was set to 4, allowing sufficent object displacement for a 10 Hz lidar. The score threshold was set to 10, which was determined experimentally on data from Town 2. We emphasize that the score threshold depends on the lidar resolution and not the application setting. For these simulated results, we used the groundtruth trajectory instead of lidar odometry to focus on the detection aspect.

\begin{figure}
    \centering
    \begin{subfigure}[b]{0.22\textwidth}
        \centering
        \includegraphics[width=\textwidth]{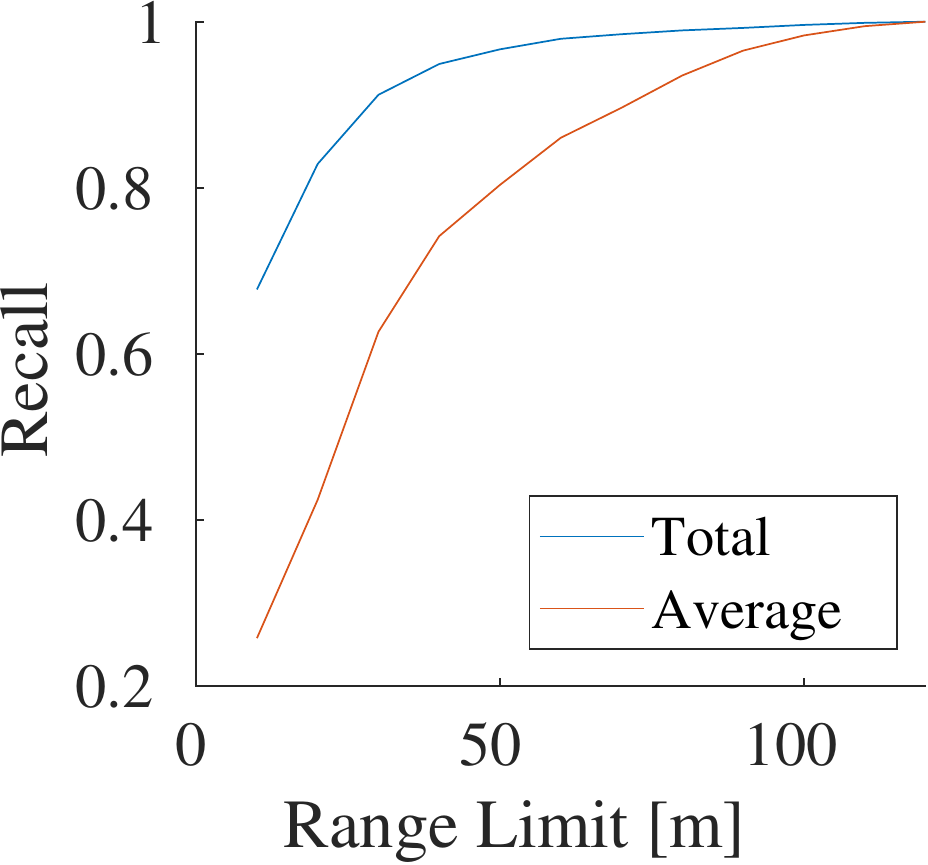}
        \caption[]%
        {{\small Town 1.}}    
        \label{fig:mean and std of net14}
    \end{subfigure}
    \quad
    \begin{subfigure}[b]{0.22\textwidth}  
        \centering 
        \includegraphics[width=\textwidth]{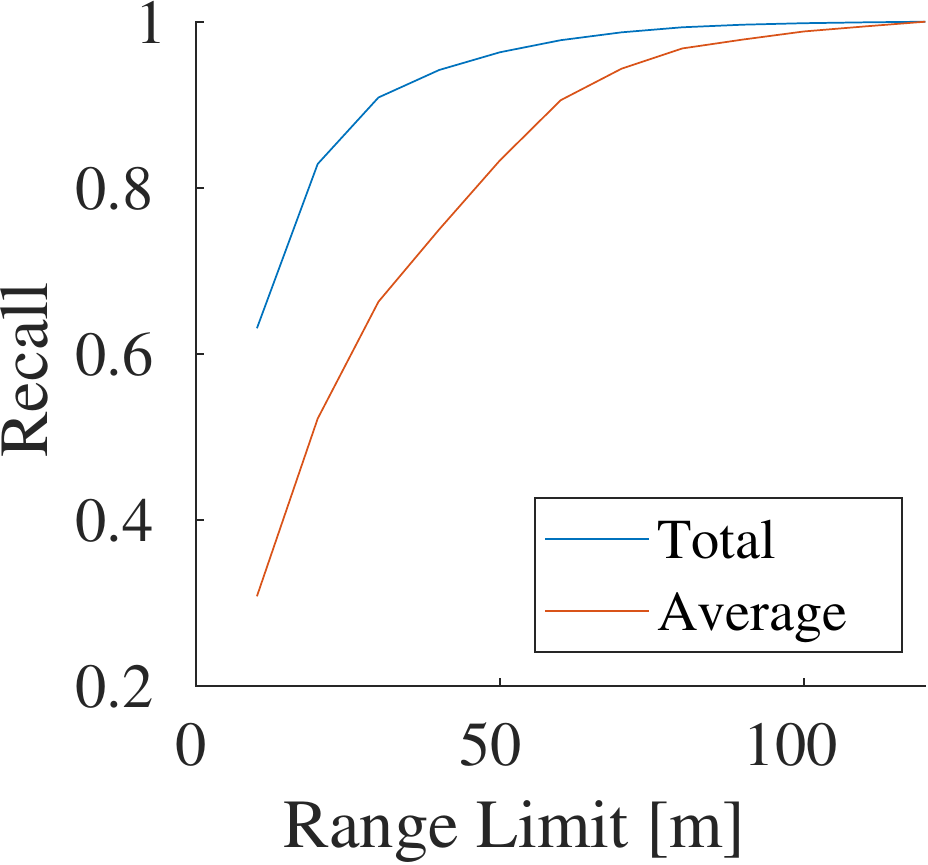}
        \caption[]%
        {{\small Town 2.}}    
        \label{fig:mean and std of net24}
    \end{subfigure}
    \caption[]
    {Recall (\textit{total} and \textit{average} - see Eq. \ref{total_pr} and \ref{average_pr}) using groundtruth labels on two simulated sequences with varying limited range. Total recall is high at low range limits because nearby objects have more points, downplaying far-away ones.} 
    \label{range}
    \vspace*{-0.6cm}
\end{figure}

\begin{figure*}[ht]
  \vspace*{0.2cm}
    \begin{subfigure}[b]{0.16\textwidth}   
        \centering 
        \includegraphics[width=\textwidth]{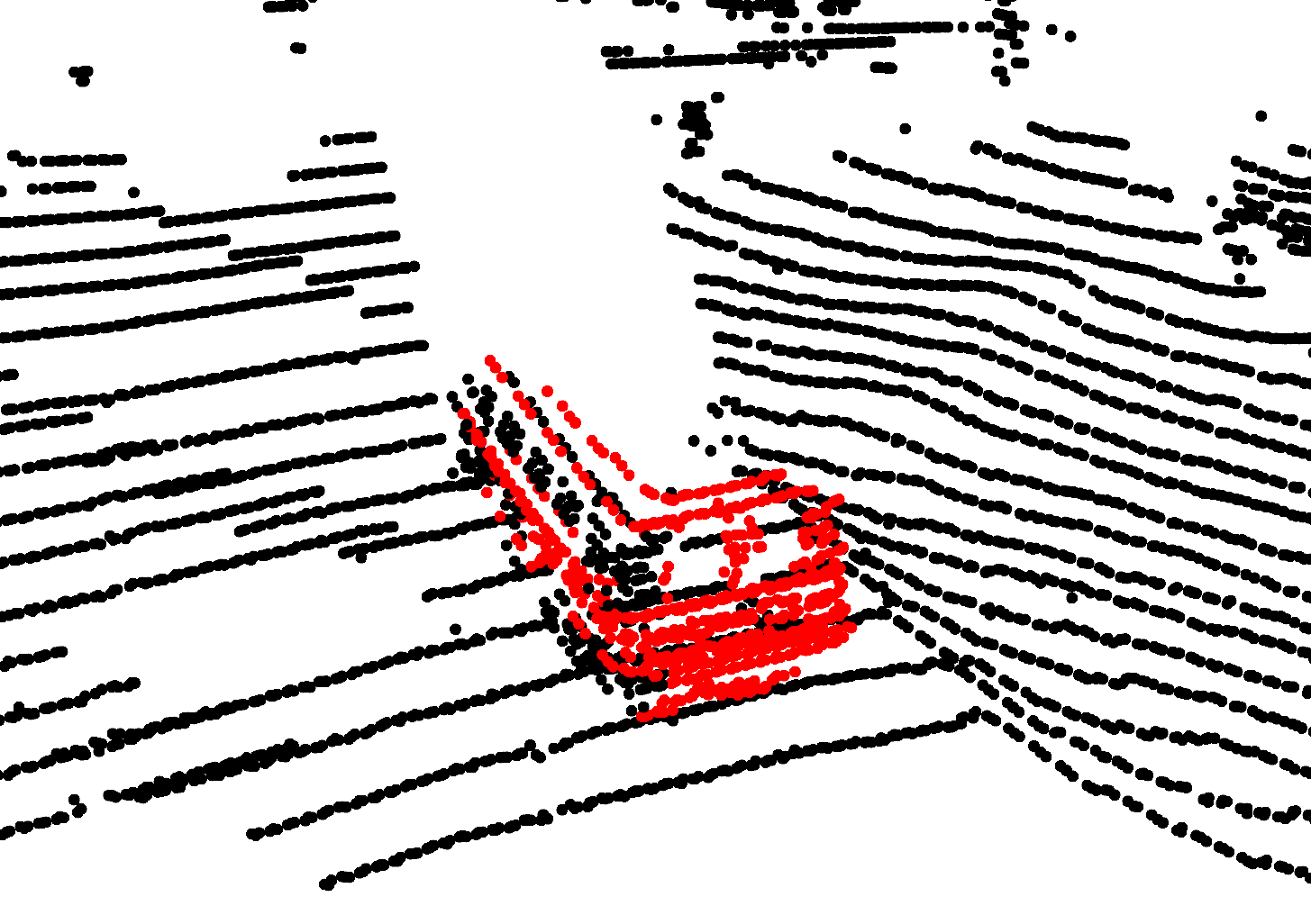}
    \end{subfigure}
    \begin{subfigure}[b]{0.16\textwidth}  
        \centering 
        \includegraphics[width=\textwidth]{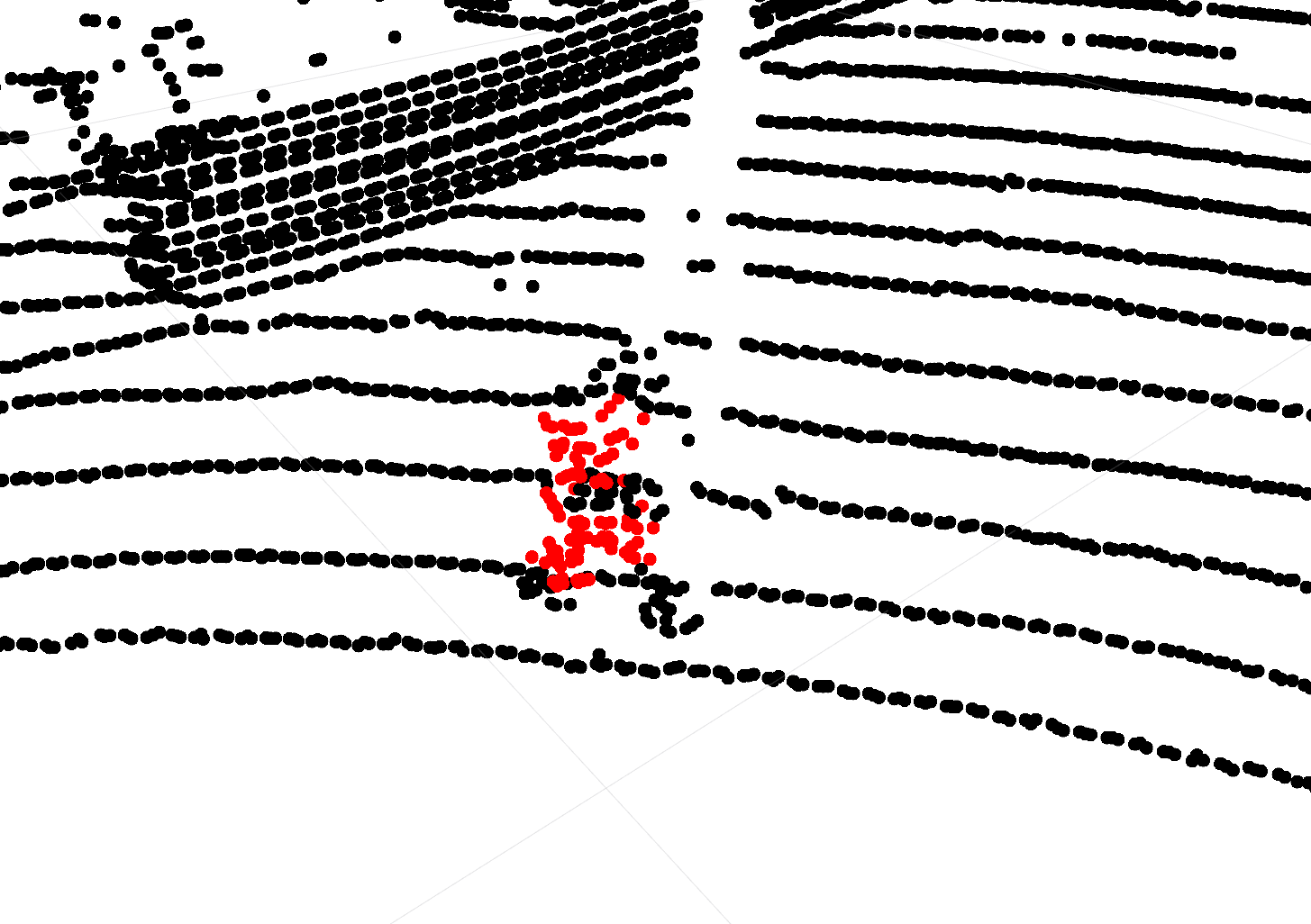}
    \end{subfigure}
    \begin{subfigure}[b]{0.16\textwidth}
        \centering
        \includegraphics[width=\textwidth]{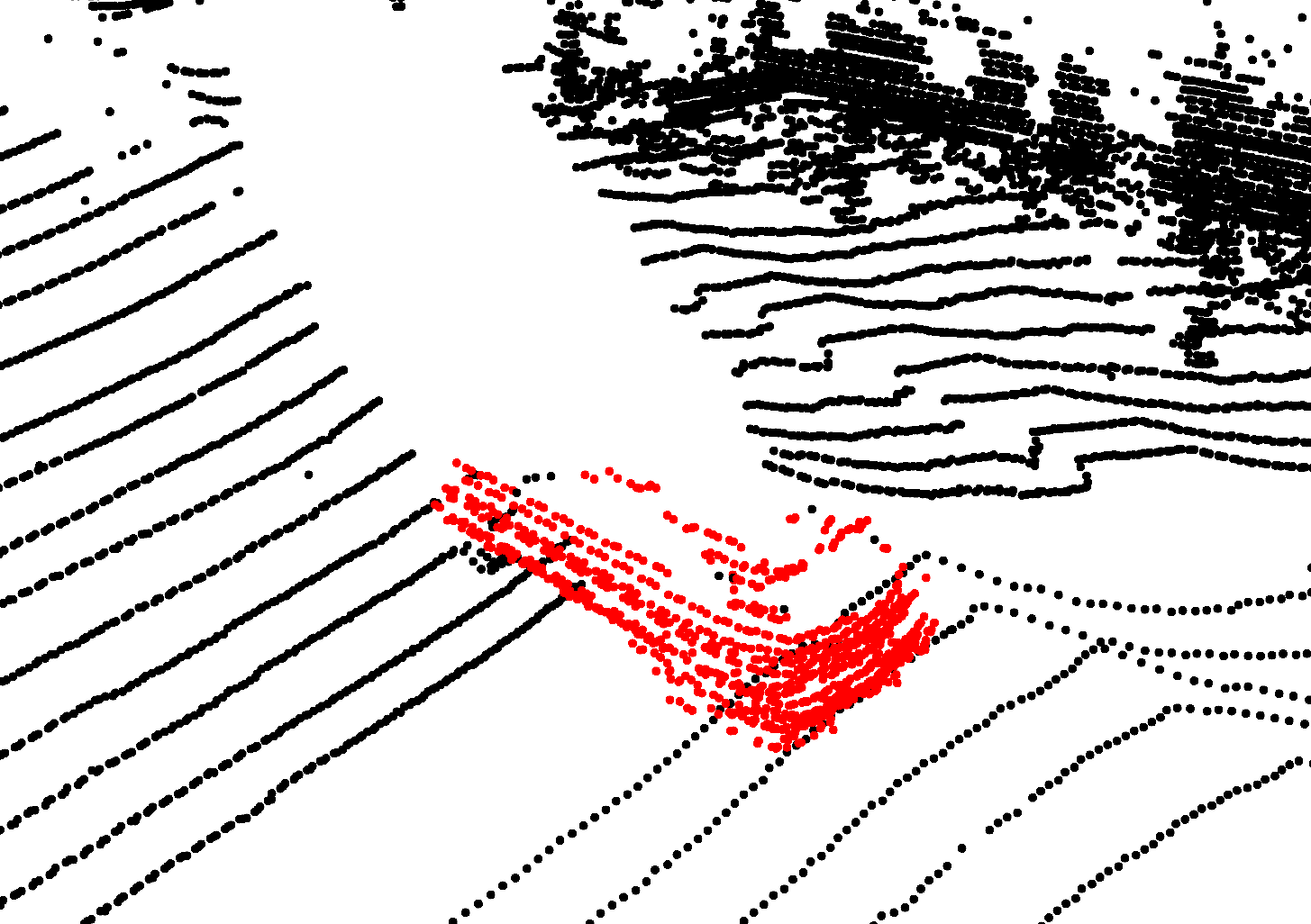}
    \end{subfigure}
    \begin{subfigure}[b]{0.16\textwidth}
        \centering
        \includegraphics[width=\textwidth]{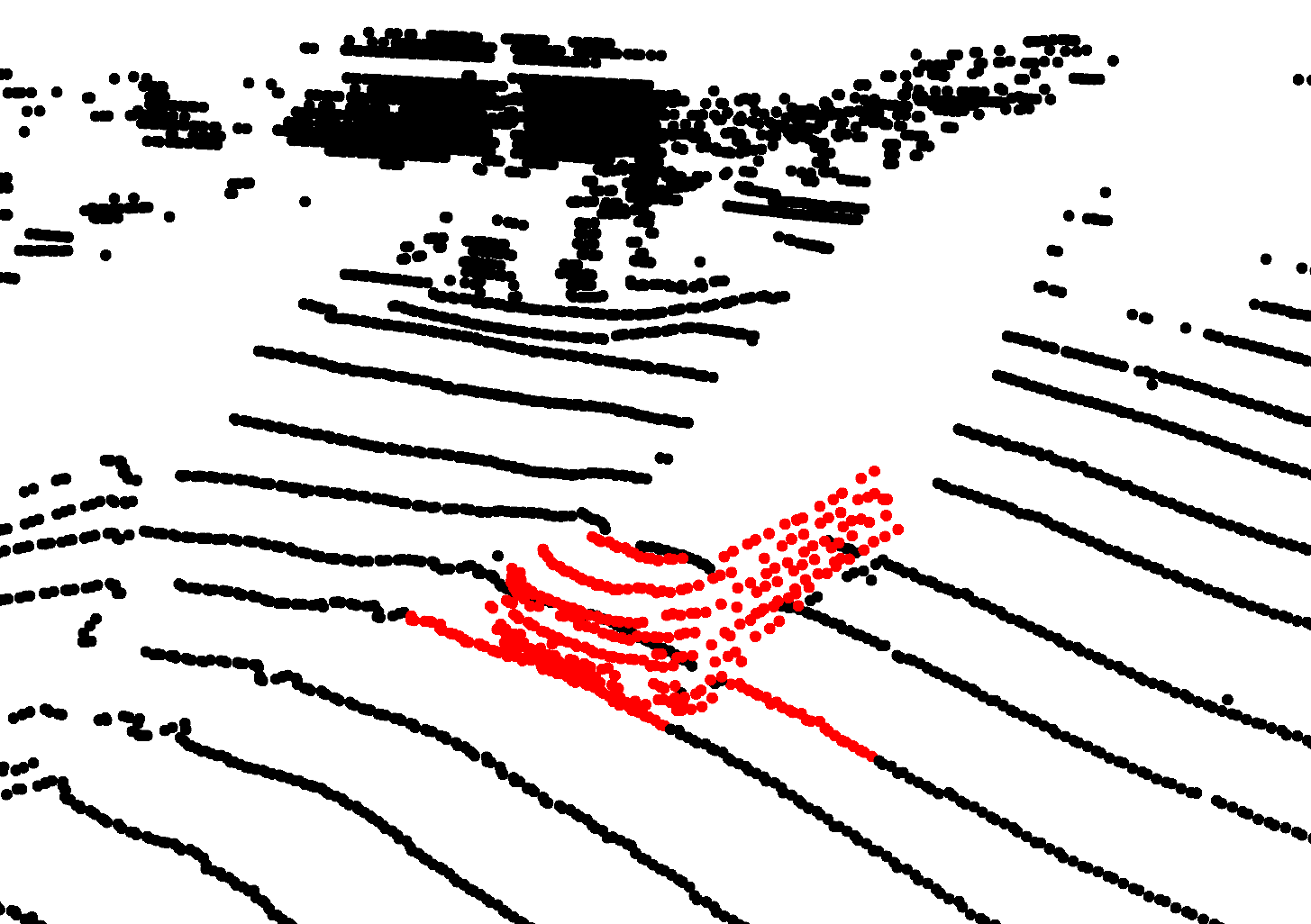}
    \end{subfigure}
    \begin{subfigure}[b]{0.16\textwidth}
        \centering
        \includegraphics[width=\textwidth]{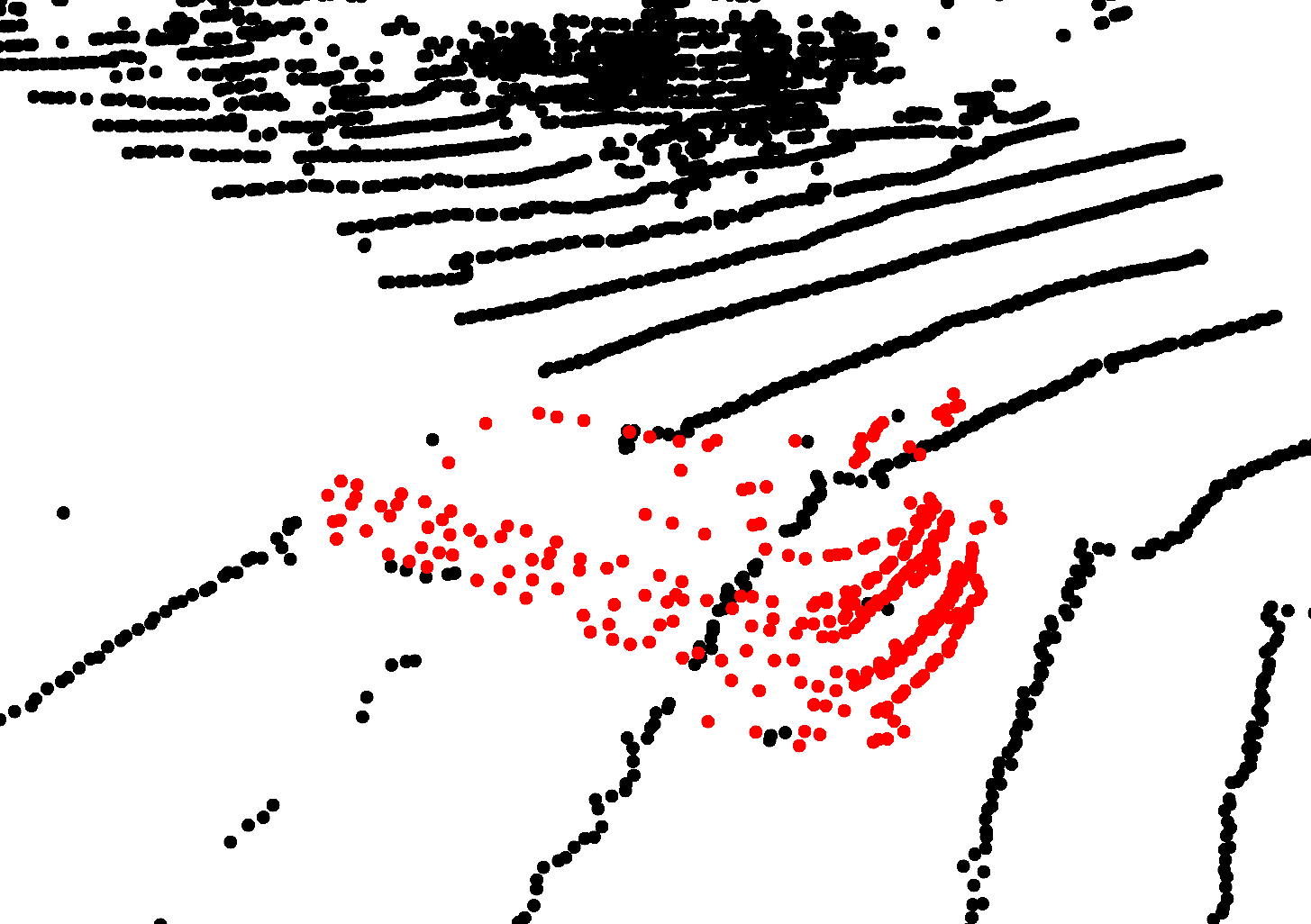}
    \end{subfigure}
    \begin{subfigure}[b]{0.16\textwidth}
        \centering
        \includegraphics[width=\textwidth]{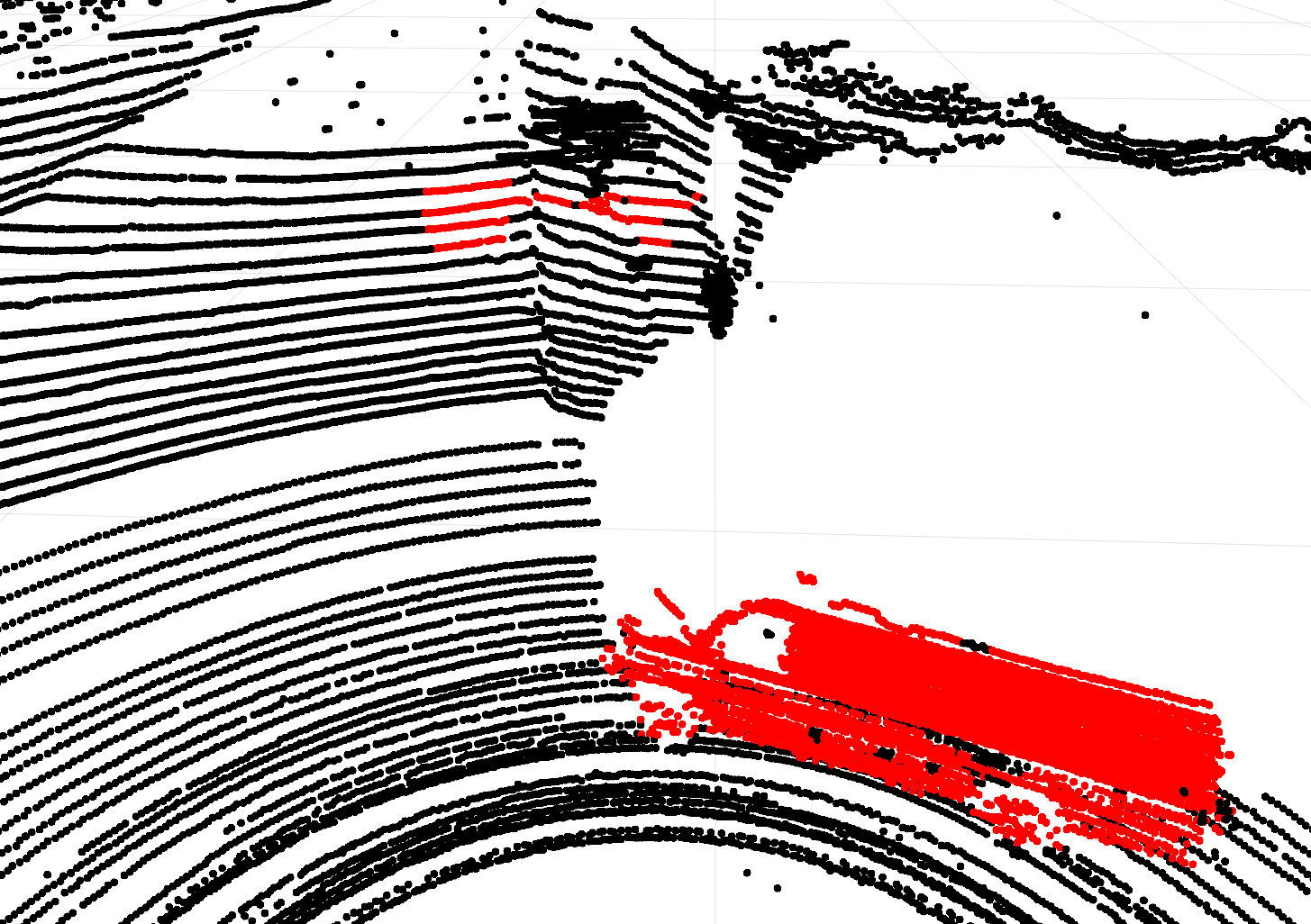}
    \end{subfigure}

    \begin{subfigure}[b]{0.16\textwidth}   
        \centering 
        \includegraphics[width=\textwidth]{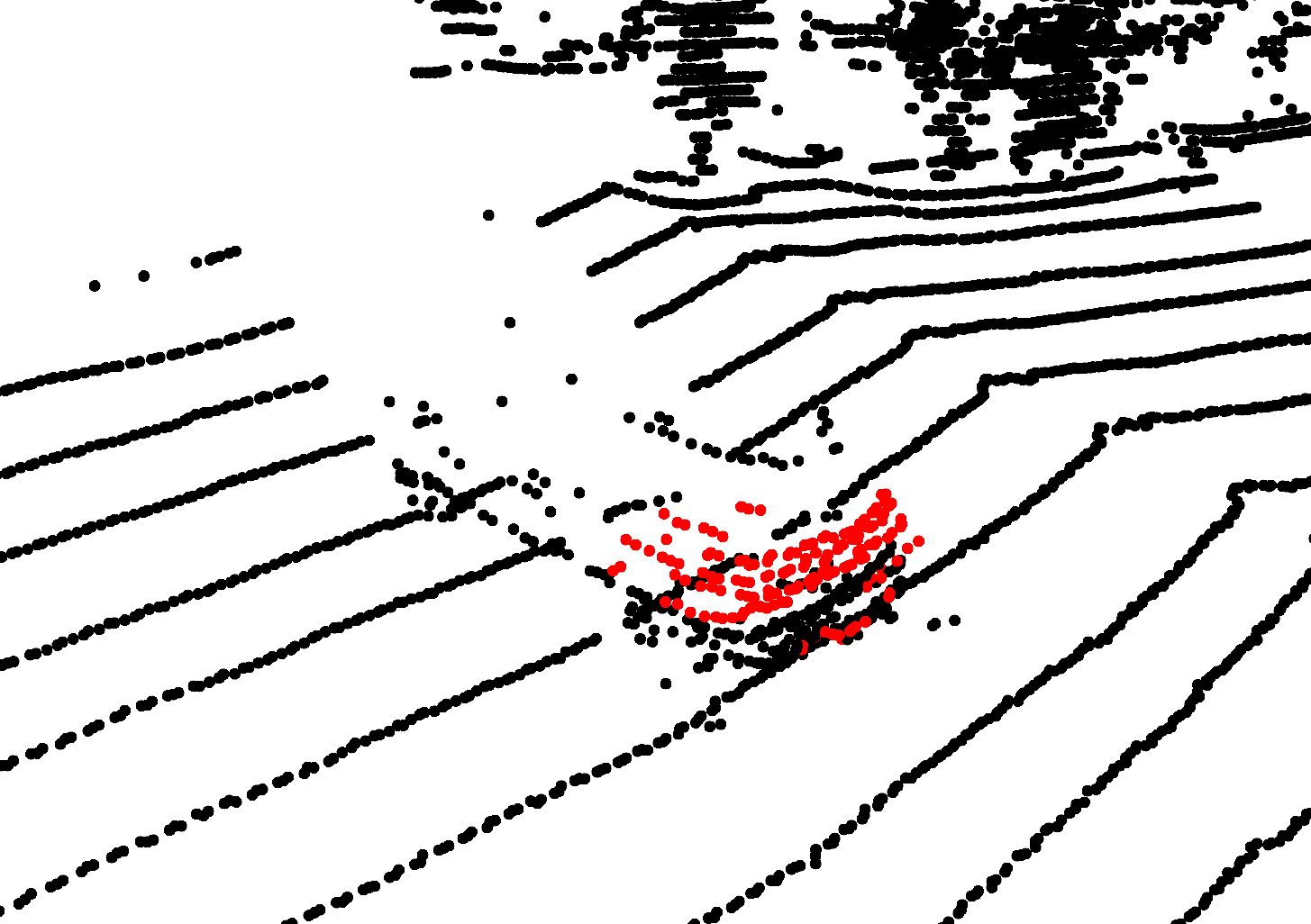}
    \end{subfigure}
    \begin{subfigure}[b]{0.16\textwidth}  
        \centering 
        \includegraphics[width=\textwidth]{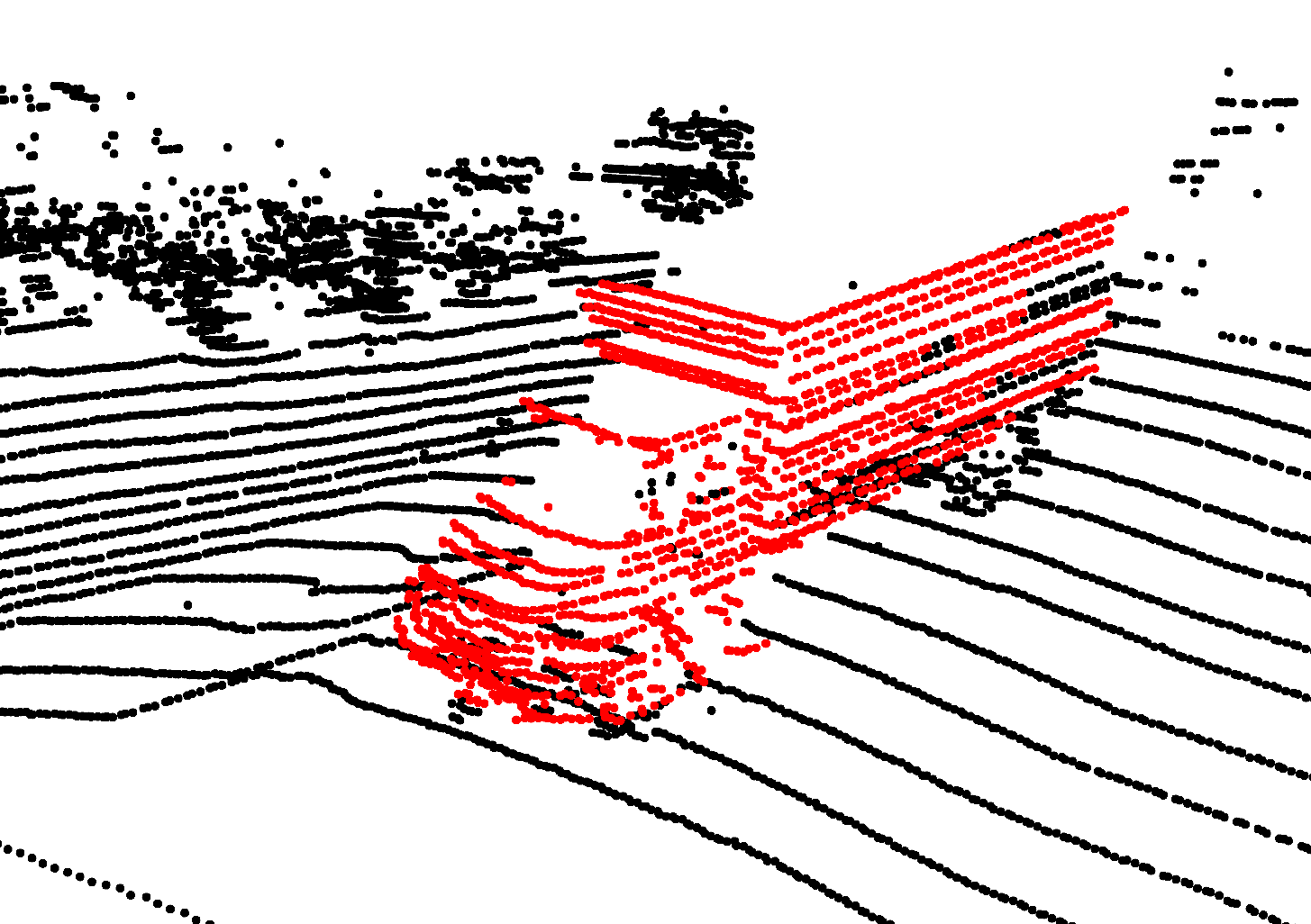}
    \end{subfigure}
    \begin{subfigure}[b]{0.16\textwidth}
        \centering
        \includegraphics[width=\textwidth]{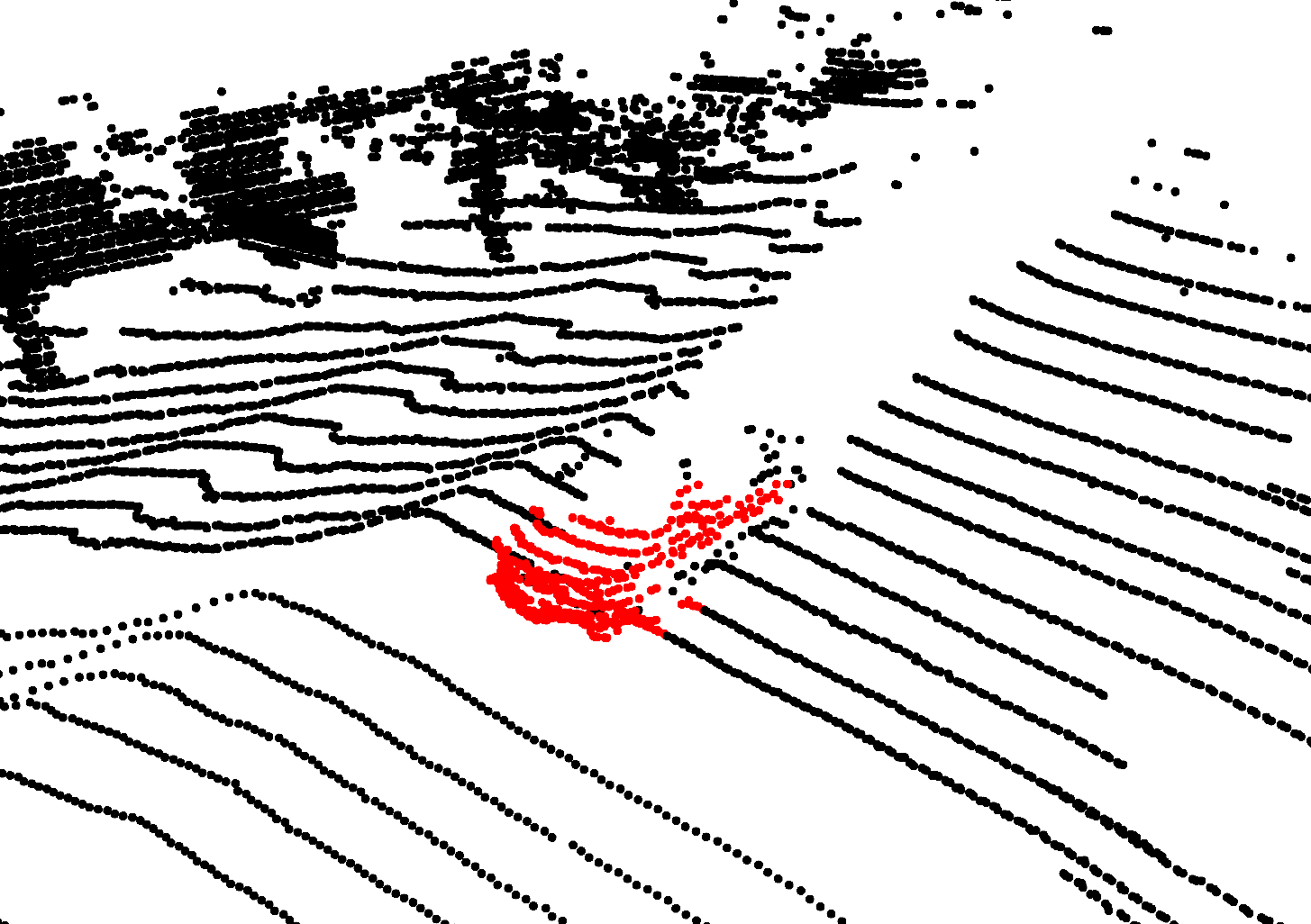}
    \end{subfigure}
    \begin{subfigure}[b]{0.16\textwidth}
        \centering
        \includegraphics[width=\textwidth]{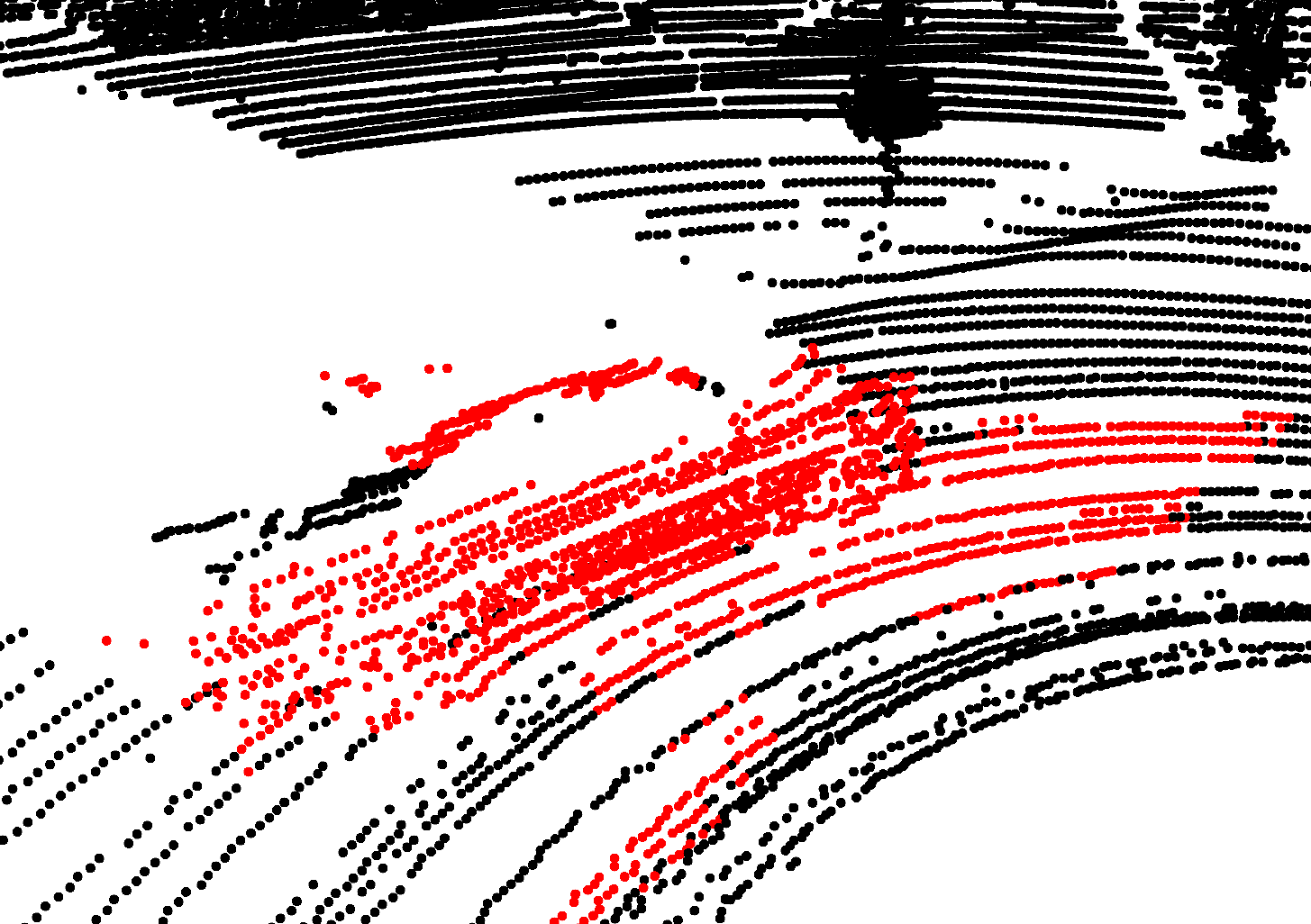}
    \end{subfigure}
    \begin{subfigure}[b]{0.16\textwidth}
        \centering
        \includegraphics[width=\textwidth]{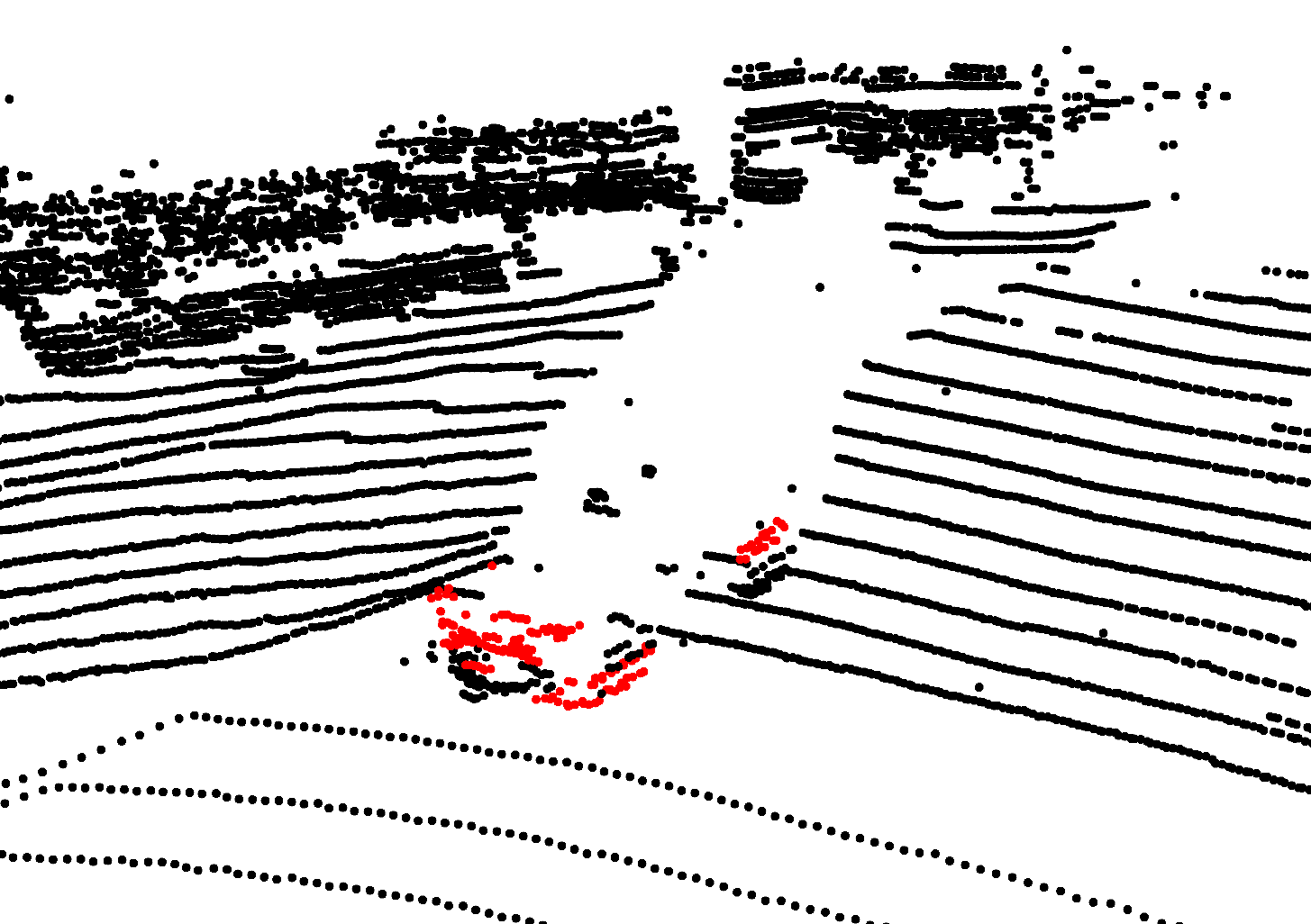}
    \end{subfigure}
    \begin{subfigure}[b]{0.16\textwidth}
        \centering
        \includegraphics[width=\textwidth]{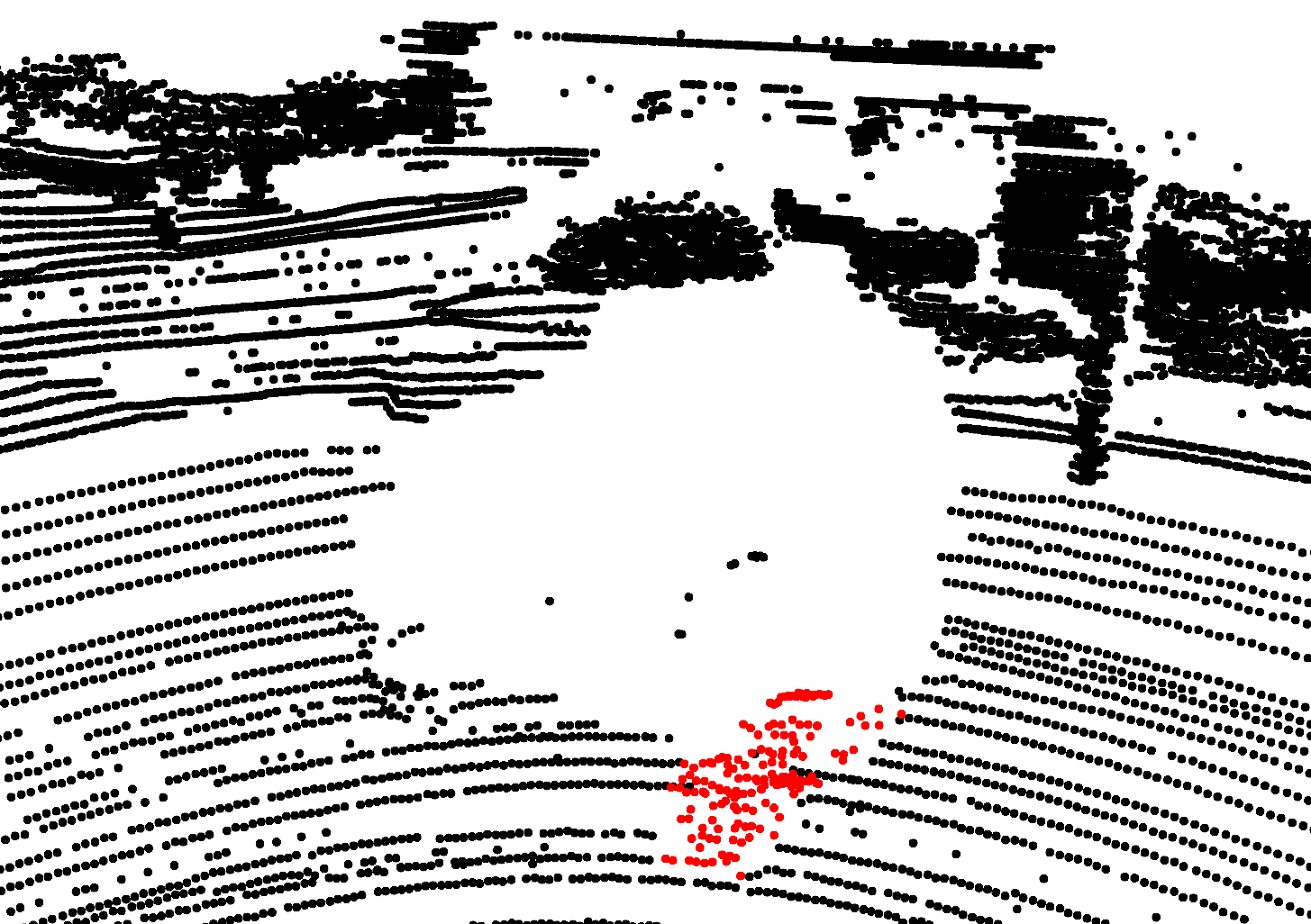}
    \end{subfigure}

    \begin{subfigure}[b]{0.16\textwidth}   
        \centering 
        \includegraphics[width=\textwidth]{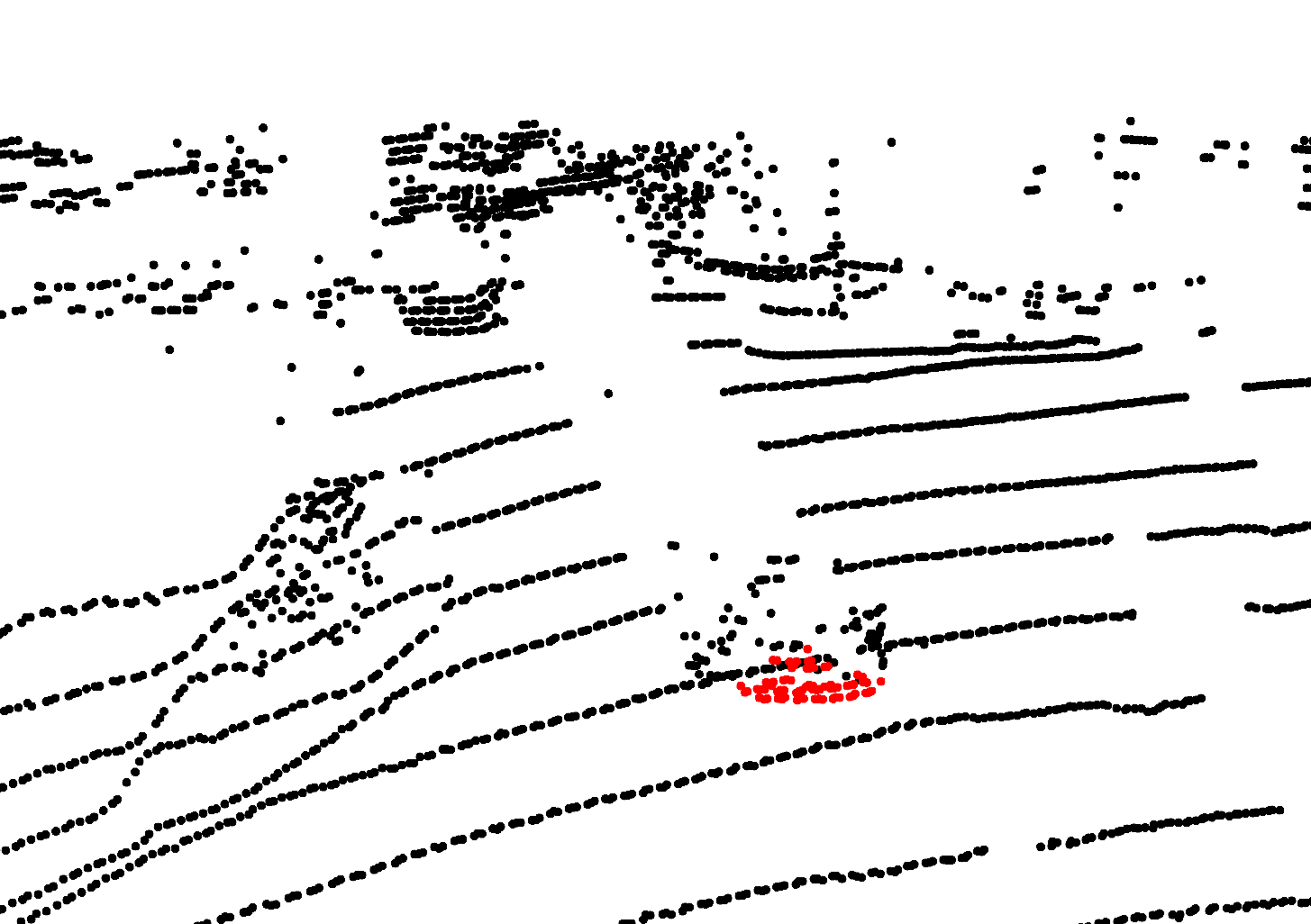}
    \end{subfigure}
    \begin{subfigure}[b]{0.16\textwidth}  
        \centering 
        \includegraphics[width=\textwidth]{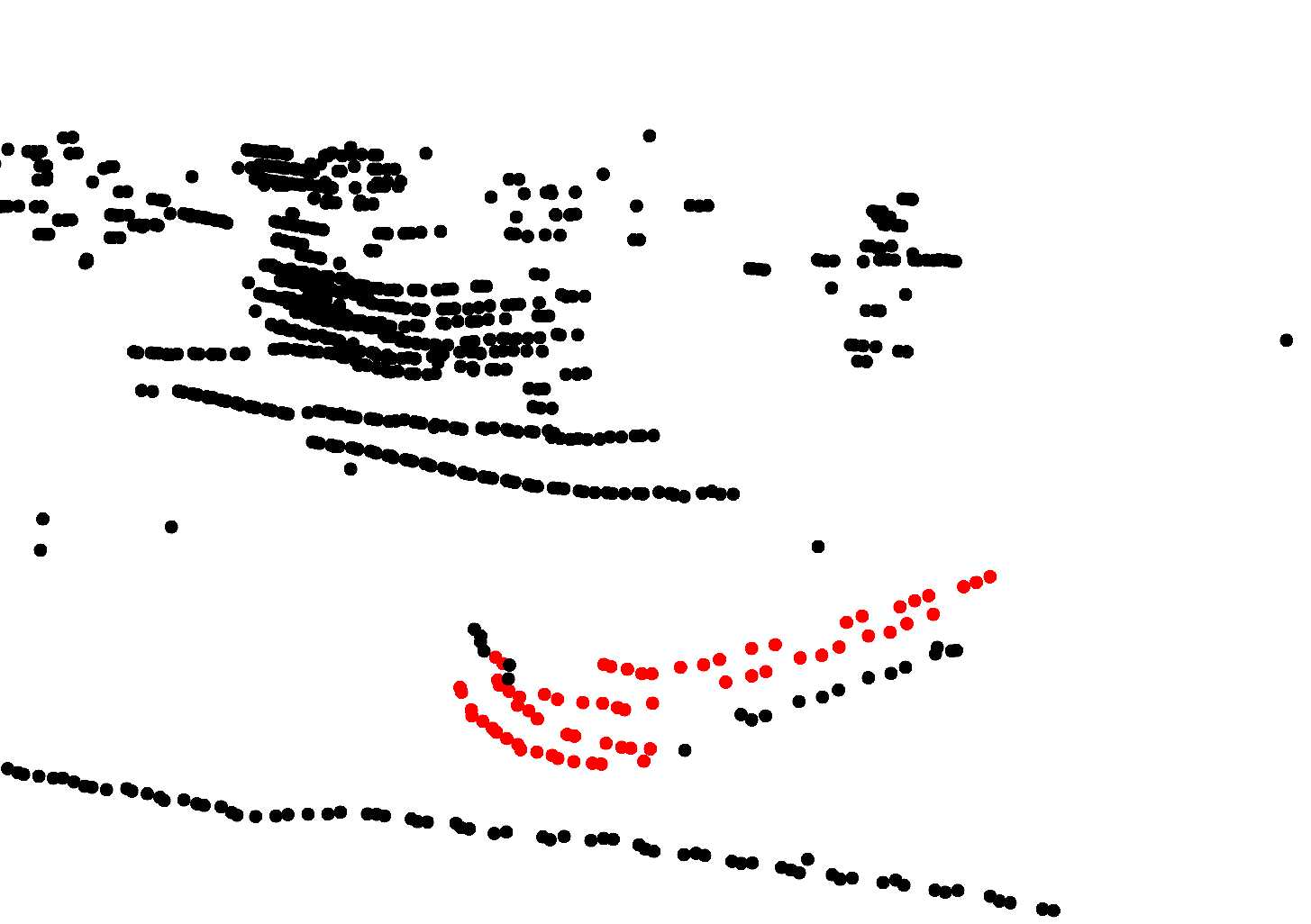}
    \end{subfigure}
    \begin{subfigure}[b]{0.16\textwidth}
        \centering
        \includegraphics[width=\textwidth]{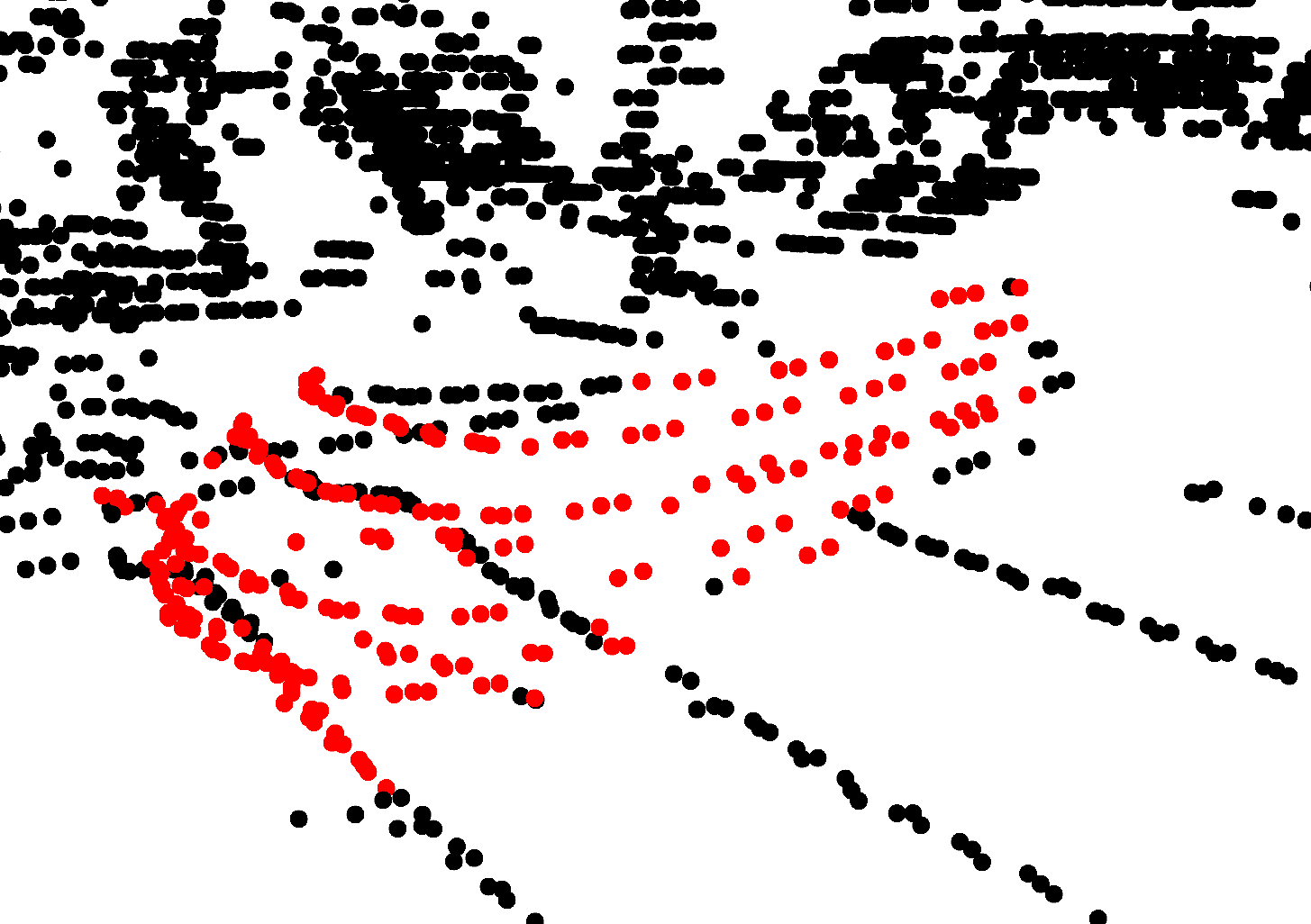}
    \end{subfigure}
    \begin{subfigure}[b]{0.16\textwidth}
        \centering
        \includegraphics[width=\textwidth]{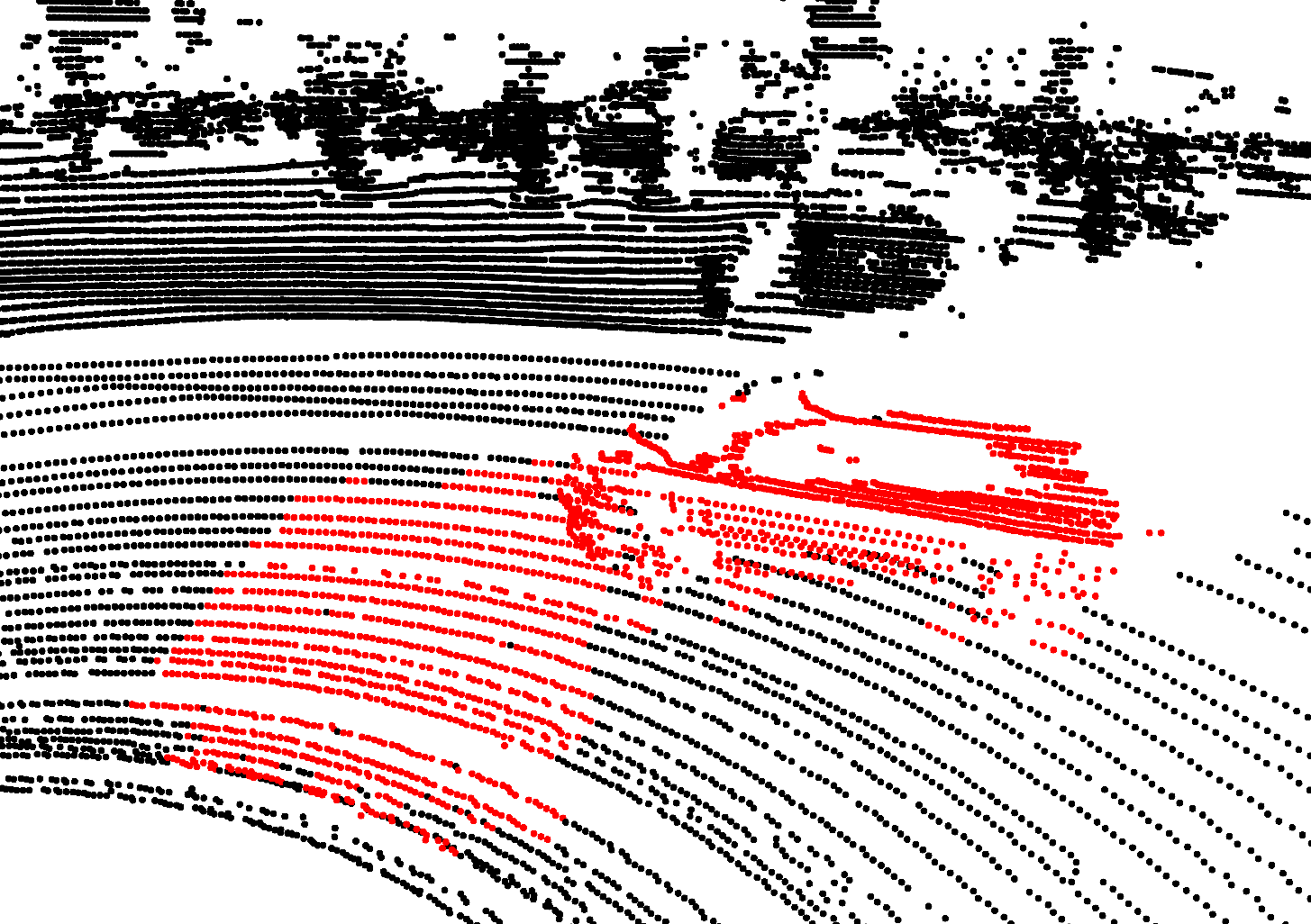}
    \end{subfigure}
    \begin{subfigure}[b]{0.16\textwidth}
        \centering
        \includegraphics[width=\textwidth]{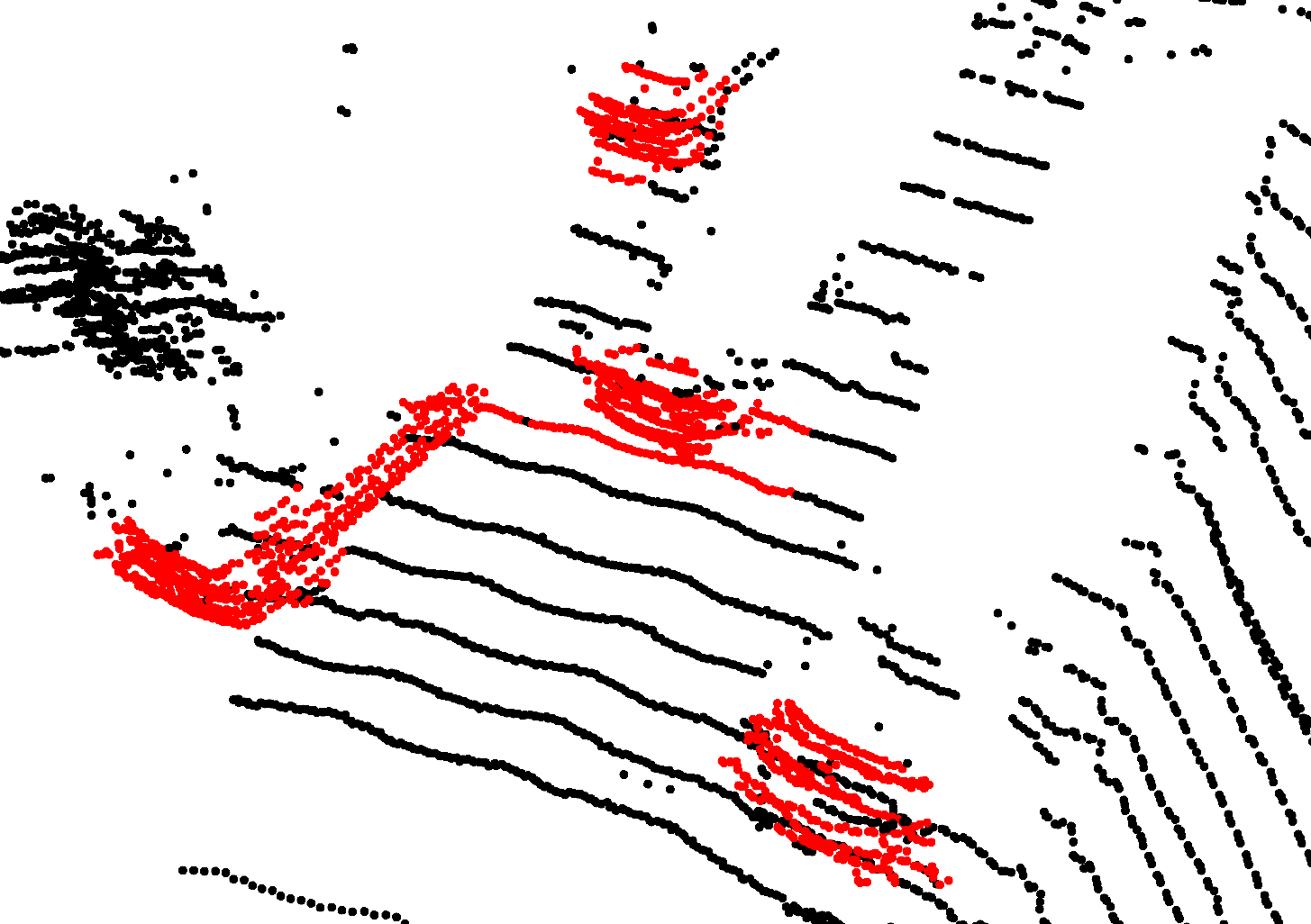}
    \end{subfigure}
    \begin{subfigure}[b]{0.16\textwidth}
        \centering
        \includegraphics[width=\textwidth]{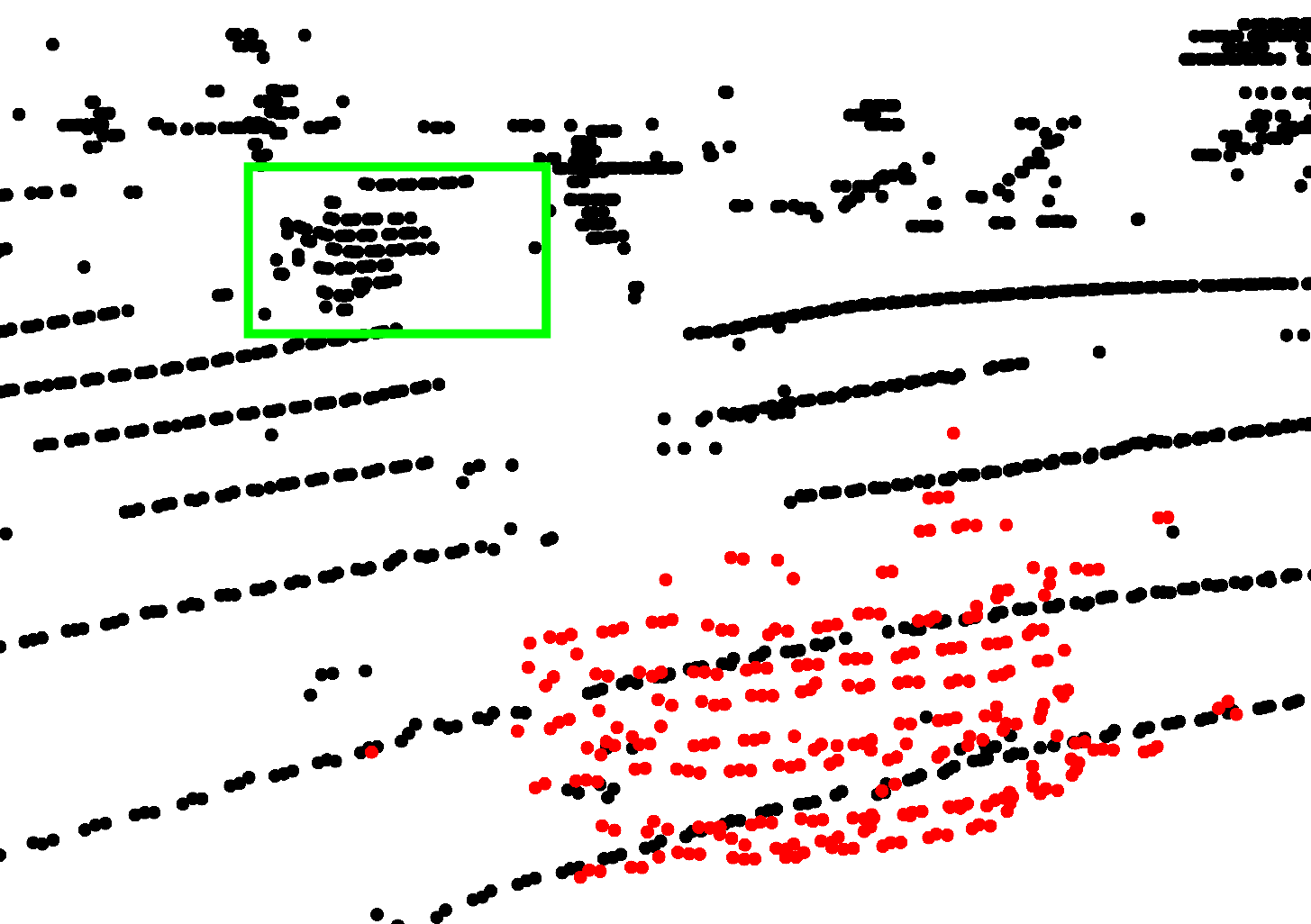}
    \end{subfigure}
    \caption[]
    {Real data examples (22 dynamic objects) of our detection method. False detections can occur (row 1, col. 6), but rarely persist in the next scan. Inaccurate surface normals cause incomplete region growth (col. 1 and 2) or growth to static points (col. 4). The green box in the last image indicates an unlabelled dynamic object due to occlusion by the other object in the freespace scan(s). The example in row 2, col. 6 is a barely visible vehicle due to many faulty returns.}
    \label{realdata}
    \vspace*{-0.4cm}
\end{figure*}

Fig. \ref{range} shows R evaluated with groundtruth at varying range limits. $\text{R}_t$ is higher for low range limits because there are more points on closer objects, downplaying the neglection of far-away objects. This also explains why range noise affects the total curves more in Fig. \ref{pr}. Greater noise reduces performance, but taking the average downplays scans with more mistakes (i.e., scans with very close objects).

We perform worse in Town 2 because of more partial occlusion instances (e.g., by fences), which we struggle with. There is latency for detecting objects accelerating from being stationary because of the scan gap, reducing R.

We hope for future comparisons to other works as our dataset is public. For now we make an indirect comparison to Dewan et al. \cite{dewan_iros17} as the state of the art. They used two manually labelled sequences of Velodyne HDL-64E data at the point level. Their pipeline without deep learning, which is setting-dependent because of ground point removal, reports the following maximum F1-score PR values: 72.8 P and 92.3 R (38 s length), and 59.5 P and 69.6 R (50 s length). Adding learning increases P at the cost of R. They do not distinguish PR in two ways like we did and is unclear about the exact computation. We stress that a fair comparison is not possible since they used real data, but we at least see our total PR values using longer, simulated sequences are comparable.

\subsection{Experiments on Real Data}
A Velodyne HDL-64E, mounted on a vehicle, was driven through Richmond Hill, Ontario. Lidar odometry was used to estimate motion. The error threshold was set to 0.5 m. Other parameters were not changed from the simulated benchmark.

A limitation was the inability to distinguish between maximum range measurements and faulty returns. Faulty returns often occur on vehicles, particularly darker ones \cite{petrov_au09}. Thus we cannot use maximum range measurements for freespace computation without incorrectly using faulty returns. This is not detrimental for ground-based applications, since Velodyne lasers are angled downward to the ground. Applications where objects have no geometry behind them for the lidar to perceive (e.g., flying objects) are an issue.

Our pipeline works well in scenarios with consistent motion (e.g., no traffic slow-downs). Fig. \ref{realdata} is a collage of real data examples, showing 22 different dynamic objects. Our pipeline struggles with occluded objects (row 3, col. 6), which is also reflected in our simulated benchmark. The pipeline also struggles with inaccurate surface normal computations, causing incomplete region growth (col. 1 and 2), or excessive growth to static points (col. 4). Row 2, col. 6 shows a barely visible vehicle due to many faulty returns.

On a laptop with an Intel Core i7-6820HQ CPU, we currently process lidar scans at 3 Hz on average on a single thread, slower than the Velodyne scan rate (10 Hz)\footnote{We intend to achieve a 10 Hz implementation before the final submission. Many individual measurement operations can be parallelized.}.

\section{CONCLUSIONS AND FUTURE WORK} \label{conclusions}
This paper presents an online detection method for labeling 3D lidar points as dynamic (moving) or static (stationary). Motion distortion is explicitly compensated, which existing methods do not consider. We only rely on the latest scans of lidar data (i.e., no maps or training data). Another trait that makes our method unique is environment isotropy. Thus our detection method is model-free and setting-independent, applicable to a wide variety of applications. We also establish and make public a benchmark with simulated motion-distorted lidar data with point-level groundtruth on dynamic objects (http://asrl.utias.utoronto.ca/datasets/mdlidar/index.html).


Future work involves resolving the issue of detecting objects without geometry behind them (e.g., flying objects), which is currently an issue because of the inability to distinguish between maximum range and faulty returns.

\section*{ACKNOWLEDGMENT}
This work was supported financially by Applanix Corporation, the Natural Sciences and Engineering Research Council (NSERC), and the Ontario Graduate Scholarship (OGS). We thank General Motors (GM) for the vehicle donation and Defence Research and Development Canada (DRDC) for the Velodyne HDL-64E loan.

\addtolength{\textheight}{-13cm}   





\newpage
\bibliography{bib/refs}

\begin{thebibliography}{10}
\providecommand{\url}[1]{#1}
\csname url@rmstyle\endcsname
\providecommand{\newblock}{\relax}
\providecommand{\bibinfo}[2]{#2}
\providecommand\BIBentrySTDinterwordspacing{\spaceskip=0pt\relax}
\providecommand\BIBentryALTinterwordstretchfactor{4}
\providecommand\BIBentryALTinterwordspacing{\spaceskip=\fontdimen2\font plus
\BIBentryALTinterwordstretchfactor\fontdimen3\font minus
  \fontdimen4\font\relax}
\providecommand\BIBforeignlanguage[2]{{%
\expandafter\ifx\csname l@#1\endcsname\relax
\typeout{** WARNING: IEEEtran.bst: No hyphenation pattern has been}%
\typeout{** loaded for the language `#1'. Using the pattern for}%
\typeout{** the default language instead.}%
\else
\language=\csname l@#1\endcsname
\fi
#2}}

\bibitem{petrov_au09}
A.~Petrovskaya and S.~Thrun, ``Model based vehicle detection and tracking for
  autonomous urban driving,'' \emph{Autonomous Robots}, vol.~26, no. 2-3, pp.
  123--139, 2009.

\bibitem{chen_cvpr17}
X.~Chen, H.~Ma, J.~Wan, B.~Li, and T.~Xia, ``Multi-view {3D} object detection
  network for autonomous driving,'' in \emph{Computer Vision and Pattern
  Recognition}, 2017.

\bibitem{hebel2011change}
M.~Hebel, M.~Arens, and U.~Stilla, ``Change detection in urban areas by direct
  comparison of multi-view and multi-temporal {ALS} data,'' in
  \emph{Photogrammetric Image Analysis}.\hskip 1em plus 0.5em minus 0.4em\relax
  Springer, 2011, pp. 185--196.

\bibitem{underwood_icra13}
J.~P. Underwood, D.~Gillsj{\"o}, T.~Bailey, and V.~Vlaskine, ``Explicit {3D}
  change detection using ray-tracing in spherical coordinates,'' in
  \emph{International Conference on Robotics and Automation (ICRA)}, 2013, pp.
  4735--4741.

\bibitem{pomerleau_icra14}
F.~Pomerleau, P.~Kr{\"u}si, F.~Colas, P.~Furgale, and R.~Siegwart, ``Long-term
  {3D} map maintenance in dynamic environments,'' in \emph{International
  Conference on Robotics and Automation (ICRA)}, 2014, pp. 3712--3719.

\bibitem{dewan_icra16}
A.~Dewan, T.~Caselitz, G.~D. Tipaldi, and W.~Burgard, ``Motion-based detection
  and tracking in {3D} lidar scans,'' in \emph{International Conference on
  Robotics and Automation (ICRA)}, 2016, pp. 4508--4513.

\bibitem{dewan_iros16}
------, ``Rigid scene flow for {3D} lidar scans,'' in \emph{Intelligent Robots
  and Systems (IROS)}, 2016, pp. 1765--1770.

\bibitem{moosmann_icra13}
F.~Moosmann and C.~Stiller, ``Joint self-localization and tracking of generic
  objects in {3D} range data,'' in \emph{International Conference on Robotics
  and Automation (ICRA)}, 2013, pp. 1146--1152.

\bibitem{ushani_icra17}
A.~K. Ushani, R.~W. Wolcott, J.~M. Walls, and R.~M. Eustice, ``A learning
  approach for real-time temporal scene flow estimation from lidar data,'' in
  \emph{International Conference on Robotics and Automation (ICRA)}, 2017, pp.
  5666--5673.

\bibitem{dewan_iros17}
A.~Dewan, G.~L. Oliveira, and W.~Burgard, ``Deep semantic classification for
  {3D} lidar data,'' in \emph{Intelligent Robots and Systems (IROS)}, 2017, pp.
  3544--3549.

\bibitem{carla}
A.~Dosovitskiy, G.~Ros, F.~Codevilla, A.~Lopez, and V.~Koltun, ``{CARLA}: {An}
  open urban driving simulator,'' in \emph{Proceedings of the 1st Annual
  Conference on Robot Learning}, 2017, pp. 1--16.

\bibitem{tang_crv18}
T.~Y. Tang, D.~J. Yoon, F.~Pomerleau, and T.~D. Barfoot, ``Learning a bias
  correction for lidar-only motion estimation,'' in \emph{Computer Robot and
  Vision (CRV)}, 2018.

\bibitem{mcgarey_fsr17}
P.~McGarey, D.~Yoon, T.~Tang, F.~Pomerleau, and T.~Barfoot, ``Field deployment
  of the tethered robotic {eXplorer} to map extremely steep terrain,'' in
  \emph{Field and Service Robotics (FSR)}, 2018, pp. 303--317.

\bibitem{azim_iv12}
A.~Azim and O.~Aycard, ``Detection, classification and tracking of moving
  objects in a {3D} environment,'' in \emph{Intelligent Vehicles Symposium
  (IV)}, 2012, pp. 802--807.

\bibitem{postica_iros16}
G.~Postica, A.~Romanoni, and M.~Matteucci, ``Robust moving objects detection in
  lidar data exploiting visual cues,'' in \emph{Intelligent Robots and Systems
  (IROS)}, 2016, pp. 1093--1098.

\bibitem{Geiger2013IJRR}
A.~Geiger, P.~Lenz, C.~Stiller, and R.~Urtasun, ``Vision meets robotics: {The}
  {KITTI} dataset,'' \emph{International Journal of Robotics Research (IJRR)},
  vol.~32, no.~11, pp. 1231--1237, 2013.

\bibitem{anderson_iros15}
S.~Anderson and T.~D. Barfoot, ``Full {STEAM} ahead: {Exactly} sparse
  {Gaussian} process regression for batch continuous-time trajectory estimation
  on {SE}(3),'' in \emph{Intelligent Robots and Systems (IROS)}, 2015, pp.
  157--164.

\bibitem{klassing_icra08}
K.~Klassing, D.~Wollherr, and M.~Buss, ``A clustering method for efficient
  segmentation of {3D} laser data,'' in \emph{International Conference on
  Robotics and Automation (ICRA)}, 2008, pp. 4043--4048.

\bibitem{moosmann_icra09}
F.~Moosmann, O.~Pink, and C.~Stiller, ``Segmentation of {3D} lidar data in
  non-flat urban environments using a local convexity criterion,'' in
  \emph{Intelligent Vehicle Symposium (IV)}, 2009, pp. 215--220.

\bibitem{roynard_ijrr18}
X.~Roynard, J.~Deschaud, and F.~Goulette, ``{Paris-Lille-3D}: {A} large and
  high-quality ground-truth urban point cloud dataset for automatic
  segmentation and classification,'' \emph{International Journal of Robotics
  Research (IJRR)}, vol.~37, no.~6, pp. 545--557, 2018.

\end{thebibliography}


\end{document}